\documentclass[10pt,twocolumn,letterpaper]{article}

\usepackage[pagenumbers]{iccv} 

\usepackage{url}
\usepackage{algorithmic}
\usepackage{algorithm}
\usepackage{array}
\usepackage{textcomp}
\usepackage{stfloats}
\usepackage{verbatim}
\usepackage{graphicx}
\usepackage{booktabs}
\usepackage{multirow}
\usepackage{makecell}
\usepackage{wrapfig}
\usepackage{pifont}
\usepackage[pagebackref,breaklinks,colorlinks,allcolors=iccvblue]{hyperref}
\newcommand{\xmark}{\ding{55}}%

\newcommand{\mypara}[1]{\noindent\textbf{#1}}

\newcommand{\name}{EVF-SAM}

\definecolor{iccvblue}{rgb}{0.21,0.49,0.74}

\title{EVF-SAM: Early Vision-Language Fusion for Text-Prompted Segment Anything Model}

\author{
    Yuxuan Zhang$^{ 1}$\thanks{Equal contribution.} \quad
    Tianheng Cheng$^{ 1*}$ \quad
    Lianghui Zhu$^{ 1}$ \quad
    Rui Hu$^{ 1}$ \quad
    Lei Liu$^{ 2}$ \quad
    Heng Liu$^{ 2}$\\
    Longjin Ran$^{ 2}$ \quad
    Xiaoxin Chen$^{ 2}$ \quad
    Wenyu Liu$^{ 1}$ \quad
    Xinggang Wang$^{ 1}$\thanks{Corresponding author (\url{xgwang@hust.edu.cn}).} \\
    $^1$ School of EIC, Huazhong University of Science \& Technology \\
    $^2$ vivo AI Lab
}

\begin{document}
\maketitle
\begin{abstract}
Segment Anything Model~(SAM) has attracted widespread attention for its superior interactive segmentation capabilities.
Despite this, the potential of text prompts for SAM remains largely unexplored.
In this paper, we investigate what text prompt encoders are beneficial for adapting SAM for Referring Expression Segmentation~(RES) and observe that (1) multimodal prompts and (2) the early-fusion mechanism incorporating text-to-image attention are crucial factors.
We introduce the \textit{\textbf{E}arly \textbf{V}ision-Language \textbf{F}usion}~(\textbf{EVF}) framework and further formulate the \textbf{\name{}}, which is simple but demonstrates superior effectiveness and efficiency compared to mainstream methods based on Large Language Models~(LLM).
We also successfully integrate segmentation data from diverse tasks into an unified hybrid dataset to conduct joint training.
The well-designed data strategies resolve issues such as semantic conflict and ambiguity.
The proposed \name{} based on BEIT-3 obtains state-of-the-art performance on RES tasks~(79.0 in average on RefCOCO/+/g), and extends RES capability to various granularities, \eg, semantic-level RES and part-level RES.
\name{} maintains the model size with only 1.32B parameters, considerably fewer than the mainstream LLM-based methods of over 7B parameters.
Code and models will be made publicly available.
\end{abstract}   
\section{Introduction}
Segment Anything Model~(SAM)~\citep{sam} brings interactive segmentation paradigm to public view. Well-trained on the SA-1B dataset, SAM achieves stunning performance and quickly becomes popular as a vision foundation model for object localization and beyond. Various SAM variants~\citep{efficientsam,mobile_sam,fast_sam,samhq} have been explored, achieving better efficiency or higher precision. Despite SAM’s surprising abilities like point-prompted and box-prompted segmentation, it is a pity that the text-prompted segmentation ability remains conceptual. We retrospect this task to Referring Expression Segmentation~(RES). 
RES focuses on the solution to predicting the segmentation mask according to the text description given by users, which enjoys several explorations by some traditional models~\citep{hu2016segmentation,liu2017recurrent,shi2018key, chen2019see,ye2019cross,hu2020bi,ding2021vision,li2021reftr,wang2022cris,lavt,polyformer,uniref++,uniseg,UNINEXT}, and is broadened by some Large Language Models~(LLM)~\citep{lisa,lisa++,pixellm,perceptiongpt,u-llava,psalm,gsva,glamm}. 

\begin{figure}[t]
    \centering
    \includegraphics[width=0.9\linewidth]{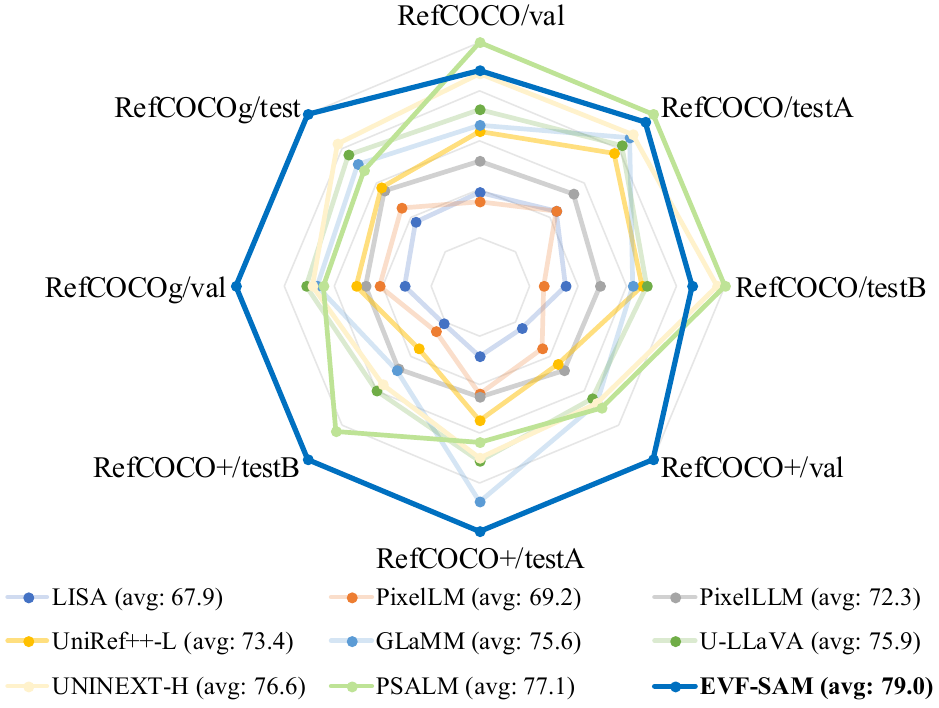}
    \vspace{-5pt}
    \caption{EVF-SAM achieves competitive performance among various benchmarks for referring expression segmentation.}
    \label{fig:radar}
    \vspace{-15pt}
\end{figure}

\begin{figure*}
    \centering
    \includegraphics[width=0.95\linewidth]{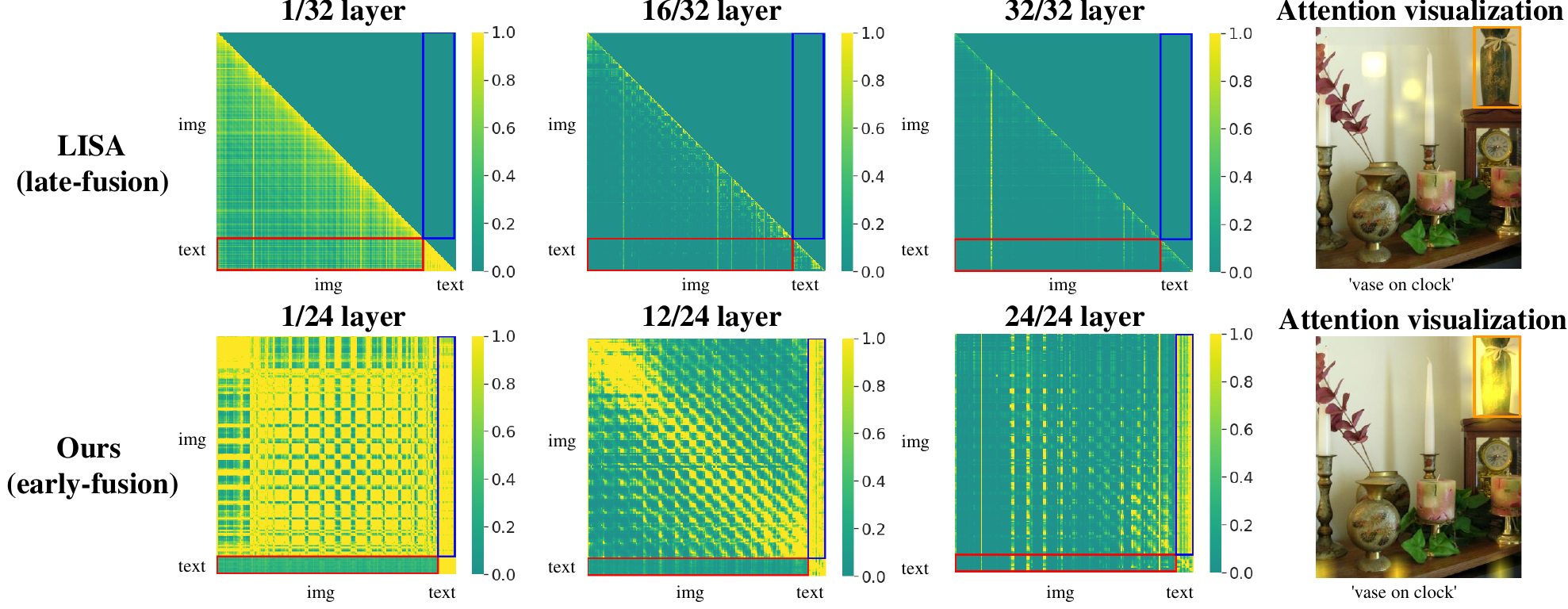}
    \vspace{-10pt}
    \caption{\textbf{Comparison between late-fusion and early-fusion.} 
    We visualize the attention map of representative late-fusion method~(\ie, LISA~\citep{lisa}) and early-fusion methods~(\ie, EVF-SAM) among different layers. The Y axis represents the `Query', the X axis represents the `Key'. We use \textcolor{red}{red}, \textcolor{blue}{blue} and \textcolor{orange}{orange} boxes to highlight the \textcolor{red}{`image to text attention'}, \textcolor{blue}{`text-to-image attention'}, and \textcolor{orange}{targeted segmentation object}. 
    The figures demonstrate that the \textcolor{blue}{`text-to-image attention'} is more crucial for Referring Expression Segmentation task, while late-fusion methods ignore it. To compensate for this, we propose the early-fusion method for high-quality text-guided visual features.
    }
    \label{fig:attention_weight}
    \vspace{-15pt}
\end{figure*}

\begin{figure*}[t]
    \centering
    \includegraphics[width=0.8\linewidth]{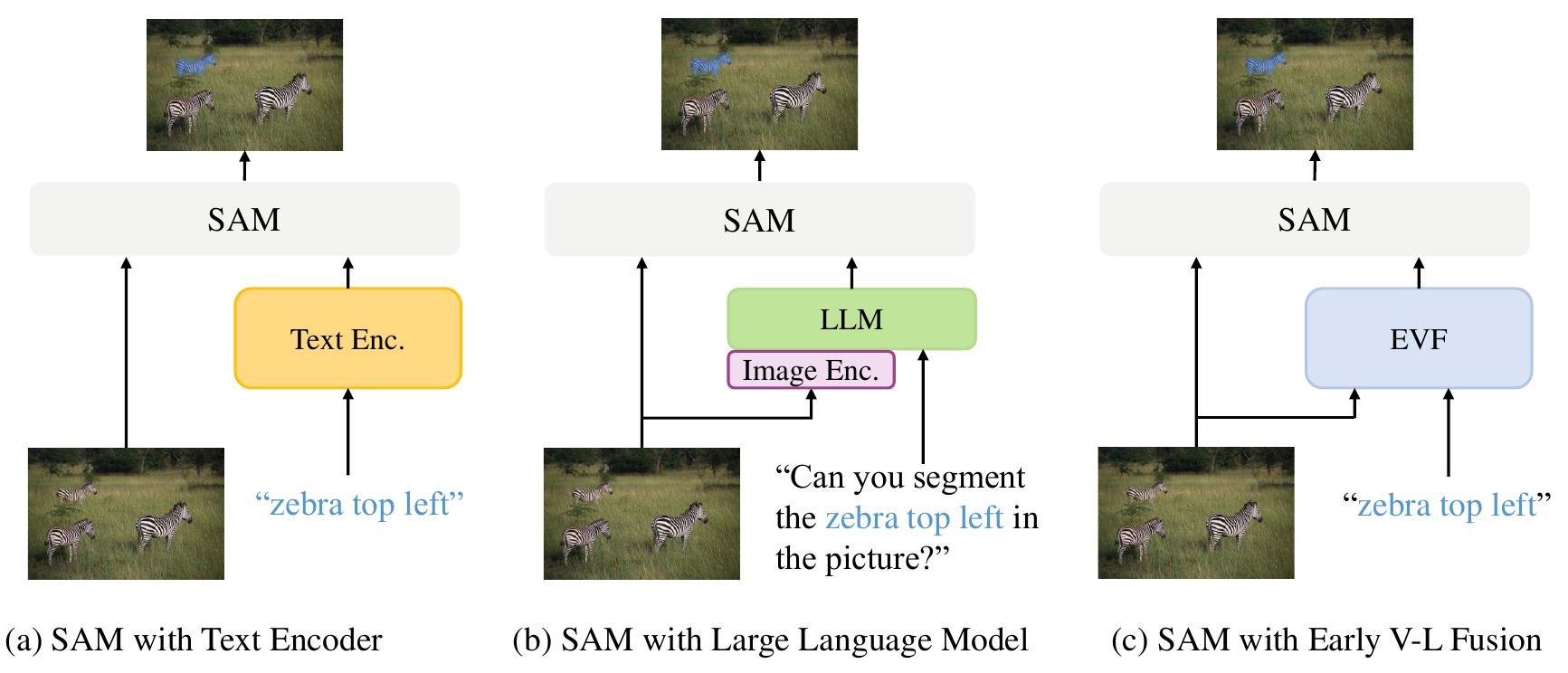}
    \vspace{-7pt}
    \caption{\textbf{Comparisons of different Text-prompted SAM.} (a) A natural idea to support text prompts is to use an \textit{off-the-shelf} text encoder to generate text embeddings for SAM~\citep{sam,refsam}. (b) Several works~\citep{lisa,lisa++,glamm} adopt Large Language Models (LLM) to generate prompt embeddings for SAM in an auto-regressive manner. (c) Our proposed \name{} exploits an effective early vision-language fusion encoder for text-prompted SAM with higher performance and fewer parameters.}
    \vspace{-13pt}
    \label{fig:comparison}
\end{figure*}

The key challenge lies in empowering SAM with language understanding ability for segmentation according to text prompts, \eg, referring expression segmentation. 
Fig.~\ref{fig:comparison} summarizes previous end-to-end works which explore the text-prompted abilities of SAM:
(a) SAM with text encoder: leverage an \textit{off-the-shelf} text encoder(\eg, CLIP~\citep{clip}) to extract text features and feed into SAM.
However, text encoders fail to extract positional information from unimodal input, leading to poor performance. 
(b) SAM with LLM: fine-tune a Large Language Model~(LLM) to auto-regressively generate embeddings for the target object. 
These methods are under the late-fusion approach, where the image features extracted by the image encoder are further integrated within the LLM. 
However, the text-to-image attention, which is proved essential~(in Fig.~\ref{fig:attention_weight}) for Referring Expression Segmentation task, is neglected in LLMs.
Additionally, these LLM-based models are often computationally expensive and hard to train.

To overcome shortcomings of existing technical approaches, we propose: (1) multimodal prompts that incorporate both text and image outperform text-only prompts; (2) the early-fusion architectures, which incorporate text-to-image attention, exhibit significant advantages over text-only encoders and Large Language Models.
Motivated by the above observations, we realize the text-promptd SAM using an \textbf{E}arly \textbf{V}ision-Language \textbf{F}usion encoder~(EVF), and propose \name{}, as shown in Fig.~\ref{fig:comparison}~(c).
The proposed \name{} is a simple yet effective framework built upon the \textit{off-the-shelf} foundation models, comprising an well-pretrained Vision-Language Model~(VLM)~(\ie, BEIT-3~\citep{beit3}), to generate prompt embeddings for SAM.
Besides, \name{} does not include elaborate designs or modules and is easy for scaling to larger models.
Moreover, \name{} reduces huge amounts of parameters~(\eg, -82\% parameters compared to LISA~\citep{lisa}), with better performance outperforming previous attempts with Large Language Models~\citep{lisa,lisa++,glamm}, as shown in Fig.~\ref{fig:radar}.

Besides the simple architecture, we highlight the importance of well-managed training data. 
We construct a hybrid dataset with publicly available datasets by including RES data~(\eg, RefCOCO/+/g~\citep{referit, refcoco/+,refcocog,refcocog2}), semantic-level data~(\eg, ADE20K~\citep{ade1, ade2}), instance-level data~(\eg, Objects365~\citep{shao2019objects365}), and part-level data~(\eg, Pascal-Part~\citep{pascal_part}). 
To effectively train the model on the diverse collection of hybrid dataset, we carefully design several data strategies for our \name{}. To be specific, we propose to use a special token `[semantic]' to solve the semantic conflict problem, and use some merging/filtering principles to solve the ambiguity problem. 
These additional data not only boost \name{} at Referring Expression Segmentation~(RES) tasks~(78.0 cIoU to 79.0 cIoU in average on RefCOCO/+/g~\citep{referit, refcoco/+,refcocog,refcocog2}), but also extend \name{}'s capabilities at various granularity~(\eg, semantic-level RES,  part-level RES). 

Our main contributions can be summarized as follows:
\begin{itemize}
    \item We propose a novel Referring Expression Segmentation (RES) framework by introducing an Early Vision-Language Fusion encoder (EVF).
    The EVF framework demonstrates superior effectiveness and efficiency compared to mainstream late-fusion frameworks, \eg, RES via Large Language Models~(LLM). 
    The EVF framework emphasizes the multimodal prompt and the early-fusion mechanism, which are crucial for the RES task, as highlighted by extensive experiments and visualizations. 
    Building upon the EVF framework, we propose \name{}.   
    \item We jointly train \name{} on a hybrid dataset, extending the model's capability to various granularities of RES tasks, \eg, semantic-level RES, part-level RES. 
    Our proposed data strategies successfully map various segmentation datasets to a unified representation, resolving distribution problems such as semantic conflict and ambiguity.
    \item \name{} achieves state-of-the-art performance on the RES tasks~(\ie, 79.0 cIoU in average on RefCOCO/+/g).
    Meanwhile, \name{} reduces the parameter count to 1.3B, a substantial decrease compared to the LLM-based models of more than 7B parameters.
\end{itemize}

\begin{figure}[t]
    \centering
    \includegraphics[width=0.9\linewidth]{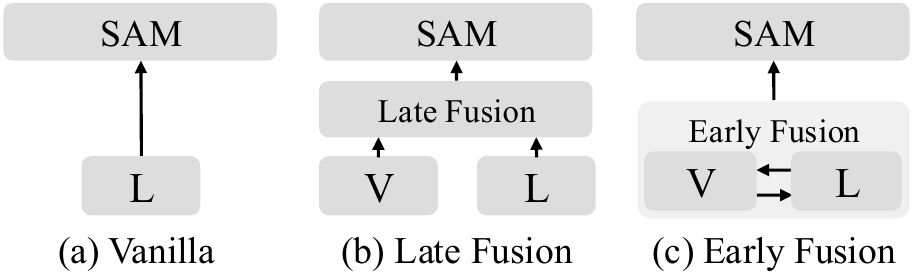}
    \vspace{-10pt}
    \caption{\textbf{Architectural explorations for text-prompted SAM.} `L' and `V' denote the text encoder and vision encoder. We mainly explore three schemes: (a) vanilla baseline with a simple text encoder, (b) multimodal inputs with a late-fusion, \ie, concatenation, and (c) multimodal inputs with early-fusion}
    \label{fig:stage_differ}
    \vspace{-19pt}
\end{figure}

\section{Related Work}

\subsection{Text-Prompted Segment Anything Models}
\label{sam_related_work}
\mypara{Segment Anything Model.} SAM~\citep{sam} is an interactive segmentation model capable of predicting masks based on various types of prompts~(points, boxes, and masks). Trained on a large-scale dataset, SAM demonstrates strong generalization capability for segmenting diverse common objects.
Advanced works address the massive computation cost of SAM or broaden SAM's capability.
Efficient-SAM~\citep{efficientsam} distills the image encoder of SAM, achieving comparable performance with significantly fewer parameters.
Fast-SAM~\citep{fast_sam}, leveraging the YOLOv8~\citep{yolov8} architecture, achieves a $50\times$ speedup for inference.
SAM-HQ~\citep{samhq} utilizes low-level features from the image encoder to address the segmentation quality of SAM.
Matte-Anything~\citep{yao2024matte} make SAM capable of outputting detailed matting masks. 
Inpaint-Anything~\citep{yu2023inpaint} integrate SAM with AIGC models to produce objects-removing tools.
VRP-SAM~\citep{sun2024vrp} enables visual prompt for referring segmentation.
Edit Everything~\citep{xie2023edit} allows users to
edit images using simple text instructions.


\mypara{Text-prompted SAM.}
Although SAM excels in visual-based segmentation tasks with box/point/mask prompts, it currently lacks language understanding abilities and it's infeasible to directly use text prompts for Referring Expression Segmentation~(RES).
Several works~\citep{groundedSAM,fast_sam,refsam} employ multi-stage architectures~(\eg, employing grounded detectors) to incorporate SAM with language interpretation capability. 
However, multi-stage models suffer from noise accumulation and difficulty in optimization, and is out of our concern.
Under end-to-end manners, vanilla SAM proposes employing a CLIP text encoder to cast linguistic feature into SAM, as shown in Fig.~\ref{fig:stage_differ} (a).
Some advanced works propose using late-fusion manner, by employing \textit{off-the-shelf} unimodal encoders to pre-extract features independently, followed by a fusion structure, as shown in Fig.~\ref{fig:stage_differ} (b).
RefSAM~\citep{refsam} employs a lightweight cross-modal MLP to project the text embeddings into SAM's prompt encoder
LISA~\citep{lisa,lisa++} employs a LLM~(LLaVA~\citep{llava}) to extract multimodal embeddings for SAM through the auto-regressive decoder.
The aforementioned methods either suffer from poor performance or are computationally expensive.
In this work, our \name{} is in `early-fusion' manner, which is both effective and efficient, as shown in Fig.~\ref{fig:stage_differ} (c).

\begin{figure*}[t]
    \centering
    \includegraphics[width=0.88\linewidth]{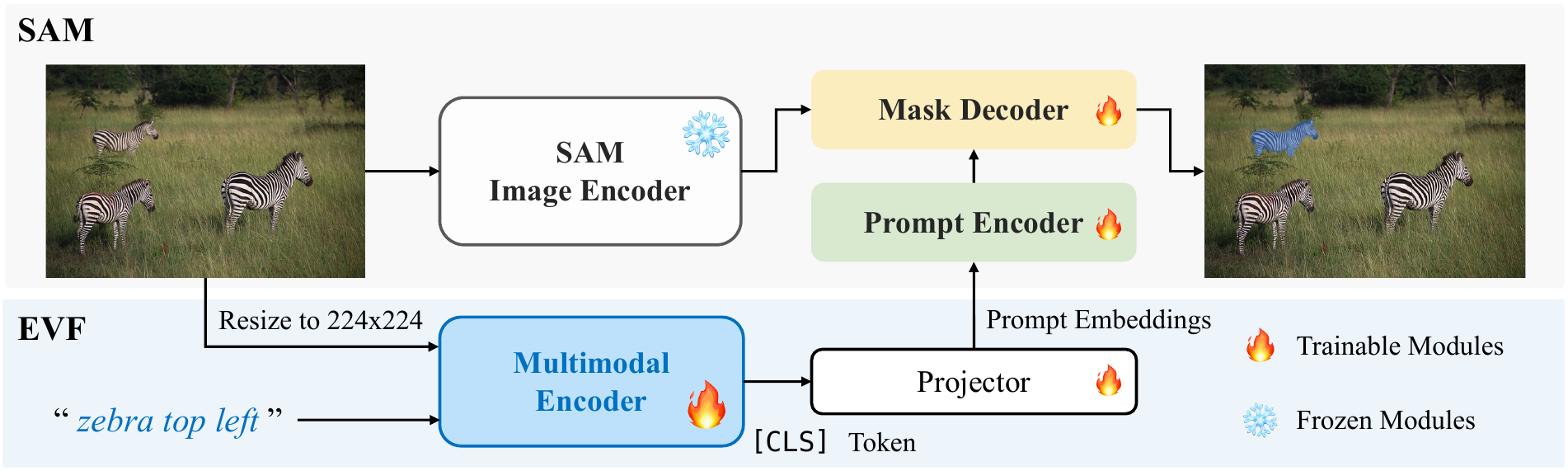}
    \vspace{-5pt}
    \caption{\textbf{The overall architecture of \name{}.} The proposed \name{} maintains the original architecture of SAM and keeps the weights of the SAM Image Encoder frozen. \name{} exploits the Multimodal Encoder with Early Vision-Language Fusion (EVF) to encode both text prompts and the low-resolution input image (which is resized to $224\times224$). Then the output \texttt{[CLS]} token is projected as prompt embeddings and fed into the prompt encoder of SAM for generating the referring segmentation results.}
    \vspace{-15pt}
    \label{fig:architecture}
\end{figure*}

\subsection{Referring Expression Segmentation}
\label{sec:res_realted_work}
Referring Expression Segmentation~(RES) is a multimodal segmentation task requiring accurate pixel-wise segmentation and fine-grained language understanding.

\mypara{RES via Text Encoders.}
Prevalent methods tend to leverage transformer-based text encoders~(\eg, BERT~\citep{bert}, CLIP~\citep{clip}) to encode texts into embeddings as guidance for segmentation. 
RefTr~\citep{li2021referring} uses a vision-language encoder to fuse multimodal features and regresses the box and mask with a carefully designed query processor.
LAVT~\citep{lavt} leverages a hierarchical ViT~\citep{vit} to perform language-aware visual encoding. 
CRIS~\citep{wang2022cris} designs a vision-language decoder to merge CLIP features, propagating fine-grained semantic information from textual representations to each pixel-level activation.
PolyFormer~\citep{polyformer} follows the encoder-decoder structure, employing a transformer decoder to generate regression results. 
Novel methods pay attention to being compatible with multiple tasks to formulate a uniform model. UNINEXT~\citep{UNINEXT}, UniRef++~\citep{uniref++} and UniLSeg~\citep{uniseg} employ similar frameworks but focus on utilizing datasets from different fields to empower their generalization capability. 
Although these traditional models are usually lightweight and achieve fine performance, they fail to integrate with large-scale foundation models(\eg, SAM\citep{sam}, LLaVA\citep{llava}), thereby struggling to keep pace with the trend of increasingly extensive pre-training.

\mypara{RES via Large Language Models.} 
In the context of the rapid development of Large Language Models~\citep{llava,Qwen-VL,emu01,emu02}~(LLM), a number of works~\citep{lisa,lisa++, pixellm, perceptiongpt, u-llava, psalm} have leveraged these models to encode expression texts for referring expression segmentation tasks.
LISA~\citep{lisa, lisa++} finetunes LLaVA~\citep{llava} to make it answer questions related to segmentation with a fixed template like `\texttt{It is [SEG]}.', where the hidden embeddings at the place of special token \texttt{[SEG]} will be seen as multimodal features. 
PixelLM~\citep{pixellm} extends LISA by building a segmentation codebook to enable multi-object segmentation. 
PixelLLM~\citep{pixelLLM} empowers the vision-language model to take locations~(\eg, a set of points or boxes) as either inputs or outputs.
PerceptionGPT~\citep{perceptiongpt} proposes an end-to-end architecture by designing a perception-enhanced vision language model, which improves inference efficiency.
u-LLaVA~\citep{u-llava} supports multi-task, \eg, object grounding.
PSALM~\citep{psalm} imports mask tokens to LLM input for better performance.
However, LLMs neglect the text-to-image attention, which is proved essential for RES in our later experiments, leading to their sub-optimal performance. Additionally, they tend to adopt heavy architectures, especially the LLMs, leading to a heavy computation burden for downstream applications.
In contrast, we find that lightweight early vision-language fusion models perform better for encode text prompts for referring image segmentation.

\subsection{Late-fusion and Early-fusion}

\mypara{Late-fusion models.}
Late-fusion denotes applying \textit{off-the-shelf} unimodal encoders(\eg, CLIP~\citep{clip}) to pre-extract features independently, followed by a fusion structure. 
This leverages the strong prior knowledge gained during the encoders' pre-training, requiring only minimal adjustment (\eg, simple projection layers) to align the multimodal feature space.
This is particularly advantageous for models with complex mechanisms~(\eg, causal attention), easing the convergence and gain the performance. 
Representative late-fusion methods are Multimodal Large Language Models~(MLLM). 
LLaVA~\citep{llava}) employs the \textit{off-the-shelf} CLIP image encoder~(CLIP~\citep{clip}) to pre-extract visual features. The visual features is projected to a Large Language Model~(\ie, LLaMA~\citep{llama}) using a simple linear layer.
Furthermore, Qwen-VL~\citep{Qwen-VL} utilizes finer-designed training recipe and carefully managed data, building a robust MLLM. Mipha~\citep{miphi} explores the lightweight architecture, following LLaVA's recipe to finetune phi~\citep{phi15}.
There are also attempts attempts to decouple causal models from late fusion, \eg, Fuyu~\citep{fuyu-8b}, EVE~\citep{EVE}. But they exhibit significantly slower convergence and poorer performance compared to late-fusion MLLMs.

\mypara{Early-fusion models.}
Early-fusion methods proposes using shallow unimodal embedding layers to project features, and then using deep fusion layers to conduct multimodal interaction.
The early integration of fusion make both modalities attach to dense information of the other one, so the extraction of multimodal feature is advantageously effective and efficient. This is crucial for tasks strongly relied on the enhance of all included modalities.
Representative early-fusion methods are Vision-Language Models~(VLM) with bidirectional attention.
ViLT~\citep{kim2021vilt} conducts detailed comparisons to prove the effectiveness and efficiency of early-fusion methods over late-fusion methods~(\eg, Pixel-BERT~\citep{huang2020pixel-bert}) in terms of multimodal feature extraction. 
BEIT-3~\citep{beit3} project visual signal to language space and performs masked language modeling pretraining, building a powerful backbone. 
ONE-PEACE\citep{one-peace} extends this approach to incorporate audio modalities, along with utilizing a larger volume of pre-training data and increasing the model's parameter size.

In this work, we proposed that `early-fusion'~(\ie, \name{}) is prioritized over `late-fusion'~(\ie, `RES via text encoder', `RES via LLMs', introduced in Sec.~\ref{sec:res_realted_work}) in the Referring Expression Segmentation task.
\begin{table*}[t]
    \centering
    \small
    \setlength{\tabcolsep}{6pt}
    \setlength{\abovecaptionskip}{0.2cm}
    \captionsetup{width=1\textwidth}
    \caption{Comparison of cIoU on different benchmarks between our proposed \name{} and previous state-of-the-art methods. \textbf{Bold}: the best results. \underline{Underline}: the second-best results. AVG represents the average metric across the eight RefCOCO-series benchmarks. We abbreviate the datasets: COCO~(C)~\citep{coco}, RefCOCO~(RC)~\citep{refcoco/+, refcocog, refcocog2, referit}, Objects365~(O)~\citep{shao2019objects365}, Video segmentation datasets~(V), ADE20K~(A)~\citep{ade1, ade2}, COCO-Stuff~(CS)~\citep{cocostuff}, PACO-LVIS~(PL)~\citep{ramanathan2023paco}, PASCAL-Part~(PP)~\citep{pascal_part}, GranD~(G)~\citep{glamm}, PASCAL VOC2010~(PV)~\citep{everingham2010pascal}, MUSE~(M)~\citep{pixellm}, gRefCOCO~(gRC)~\citep{liu2023gres}, COCO-Interactive~(CI)~\citep{psalm}, FSS-1000~(F)~\citep{li2020fss}, SA-1B~(SA)~\citep{sam}, PartImageNet~(PIN)~\citep{he2022partimagenet}, HumanParsing (HP)~\citep{liang2015human,liang2015deep}, GoldG (GG)~\citep{GoldG}.}
    \vspace{0pt}
    \scalebox{0.85}{
    \newcommand{\tbb}[1]{\textcolor{blue}{#1}}
    \begin{tabular}{llclccccccccc}
         \toprule
         \multirow{2}{*}{Method} & \multirow{2}{*}{\makecell[c]{Text Prompt\\Encoder}} & \multirow{2}{*}{{SAM?}} & \multirow{2}{*}{Training Data} &\multicolumn{3}{c}{RefCOCO} & \multicolumn{3}{c}{RefCOCO+} & \multicolumn{2}{c}{RefCOCOg} & \multirow{2}{*}{AVG}  \\\cmidrule{5-12}
         &&&& val  & testA & testB & val & testA & testB & val & test &  \\
         \midrule
         LAVT~$_\text{\citep{lavt}}$ & BERT-B~(104M) & \xmark & RC, gRC & 72.7 & 75.8 & 68.8 & 62.1 & 68.4 & 55.1 & - & - & -\\
         PolyFormer-L~$_\text{\citep{polyformer}}$ & BERT-B~(104M) & \xmark & RC, gRC & 76.9 & 78.5 & 74.8 & 72.2 & 75.7 & 66.7 & 71.2 & 71.2 & 73.4\\
         UNINEXT-H~$_\text{\citep{UNINEXT}}$ & BERT-B~(104M) & \xmark & O, C, RC, V & 82.2 & 83.4  & \underline{81.3}  & 72.5 &	76.4  &	66.2  &	74.4 & 76.4 & 76.6\\
         UniLSeg-100~$_\text{\citep{uniseg}}$ & CLIP-B~(63M)  & \xmark & SA, RC, gRC & 81.7 & 83.2 & 79.9  & 73.2 & 78.3  &	68.2  &	-    &	-   & -\\
         UniRef++-L~$_\text{\citep{uniref++}}$ & BERT-B~(104M) &\xmark & RC, F, V  & 79.1 & 82.1  & 77.5  & 68.4 &  74.0  & 61.5  & 71.4 & 72.8 & 73.4 \\
         \midrule
         LISA~$_\text{\citep{lisa}}$ & Vicuna~(7B) & \checkmark & A, CS, RC, PL, PP & 74.1 & 76.5  & 71.1  & 62.4 &	67.4  &	56.5  & 66.4 & 68.5 & 67.9 \\
         PixelLM~$_\text{\citep{pixellm}}$ & LLaMA2~(13B) & \xmark &A, CS, RC, PL, M&  73.0 & 76.5  & 68.2  & 66.3 &	71.7  &	58.3  &	69.3 & 70.5 & 69.2 \\
         PixelLLM~$_\text{\citep{pixelLLM}}$ & T5-XL~(3B) & \checkmark & RC, GG & 76.9 & 78.5 & 74.4 & 69.2 & 72.1 & 64.5 & 70.7 & 72.4 & 72.3\\
         GLaMM~$_\text{\citep{glamm}}$  & Vicuna~(7B)  & \checkmark & G, RC & 79.5 & 83.2  & 76.9  & 72.6 &	\underline{78.7}  &	64.6  &	74.2 & 74.9 & 75.6\\
         u-LLaVA~$_\text{\citep{u-llava}}$ & Vicuna~(7B) & \checkmark &A, CS, RC, PL, PV & 80.4 & 82.7 & 77.8 & 72.2 & 76.6 & 66.8 & 74.8 & 75.6  & 75.9 \\
         PSALM~$_\text{\citep{psalm}}$ & Phi-1.5~(1.3B) & \xmark & C, RC, CI & \textbf{83.6} & \textbf{84.7}  & \textbf{81.6}  & 72.9 &	75.5  &	\underline{70.1}  &	73.8 & 74.4 & 77.1 \\
         \midrule
         \name{} & BEIT-3~(673M) & \checkmark & RC& 82.1 & 83.7  & 80.0  & \underline{75.2} & 78.3 &	\underline{70.1}  &	\underline{76.8} & \underline{77.4} & \underline{78.0}\\
         \name{} & BEIT-3~(673M) & \checkmark & RC, O, A, PP, PIN, HP & \underline{82.4} & \underline{84.2} & 80.2 & \textbf{76.5} & \textbf{80.0} & \textbf{71.9} & \textbf{78.2} & \textbf{78.3} & \textbf{79.0} \\
         \bottomrule
    \end{tabular}}   
    \label{tab:benchmark}
    \vspace{-15pt}
\end{table*}

\section{Method}

\subsection{Motivation: Early Vision-Language Fusion.}
\label{motivation-arch}

We investigate how to encode text prompts for SAM in this section.
We start by using a text encoder following SAM's trial, and extend to late-fusion and early-fusion, as shown in Fig.~\ref{fig:comparison}. We conduct preliminary experiments on RefCOCO (testA), shown in Tab.~\ref{tab:mm_prompt}.


\begin{table}[t]
    \centering
    \setlength{\abovecaptionskip}{0.2cm}
    \small
    \caption{\textbf{Motivation analysis.} Both CLIP and BEIT-3 are of \texttt{Large} scale, with comparable numbers of parameters. Specifically, CLIP has a total parameter count of 428M, while BEIT-3 totals 673M parameters. `L' and `V' denote the linguistic input and the vision input. The reported metrics are evaluated on RefCOCO/testA, except that the metric of LLaVA~\citep{llava} is borrowed from LISA-7B~\citep{lisa}.}
    \vspace{2pt}
    \begin{tabular}{rccccc}
    \toprule
    & \makecell{CLIP\\(L)} & \makecell{CLIP\\(L+V)} & \makecell{BEIT-3\\(L)} & \makecell{BEIT-3\\(L+V)} & \makecell{LLaVA\\(L+V)} \\
    \midrule
    cIoU & 63.4 & 67.9 & 65.1 & \textbf{83.7} & 79.1\\
    \bottomrule
    \end{tabular}
    \vspace{-15pt}
    \label{tab:mm_prompt}
\end{table}

\mypara{Multimodal referring information is crucial.}
The interactive capability of SAM~\citep{sam} hinges on its prompt encoder, which converts points, boxes, and masks into embeddings that incorporate positional information.
As a result, when we are dealing with the Referring Expression Segmentation~(RES) task based on SAM, the key challenge is to get the positional information from the input referring prompt. 
SAM~\citep{sam} explored text-prompted segmentation with a CLIP text encoder, as illustrated in Fig.~\ref{fig:comparison}~(a). However, it is infeasible for the unimodal text encoder to extract positional information. Our reproduced metrics in Tab.~\ref{tab:mm_prompt} shows that CLIP-prompted SAM achieves 63.4 cIoU, which is inferior to well-defined baselines.
Moreover, we observe performance improvements after using multimodal prompts, \ie, 63.4 \textit{v.s.} 67.9 for CLIP and 65.1 \textit{v.s.} 83.7 for BEIT-3, proving that multimodal referring information is effective for text-prompted SAM.

\mypara{Early-fusion outperforms late-fusion.}
Existing works employ pre-encoded text feature~(\eg, RES via text encoder) or pre-encoded image feature~(\eg, RES via LLM), as illustrated in Sec.~\ref{sec:res_realted_work}, demonstrating the `late-fusion' manner. 
We visualize the attention map within multimodal fusion layers in Fig.~\ref{fig:attention_weight}, and find shortcomings within `late-fusion' methods. The attention map $\textbf{W}$ is calculated by $\textbf{W}=\text{clamp}_0^1(2\times\text{sigmoid}(QK^{\mathrm{T}}/\sqrt{d}))$, where $Q$ denotes query, $K$ denotes Key, and $\text{clamp}(\cdot)$ is a clipping function for a more distinct comparison
We firstly observe enhanced image-to-text attention at the first few layers, indicating that `RES via text encoder' methods, which lack image-to-text attention, may struggle with accurately interpreting text descriptions.
Moreover, the `text-to-image attention' takes the lead at later layers, showing the inefficiency of `RES via LLM' methods, which results in a limited multimodal perception of visual features.
The attention visualization example in Fig.~\ref{fig:attention_weight} further illustrates the imprecise object localization capabilities of late-fusion methods.
In contrast, our proposed `early-fusion' methods which utilize shallow embedding layers and deep fusion layers (\eg, \citep{kim2021vilt,beit3}) is effective and efficient  at both `image-to-text attention' and `text-to-image attention', overcoming the shortcomings of `late-fusion'.
Experiment metrics in Tab.~\ref{motivation-arch} further support our motivation. 
SAM with CLIP-text and CLIP-image, which implements a straightforward `late-fusion' approach through concatenation, achieves a cIoU of 67.9. 
Furthermore, SAM with LLaVA, representing a more complex form of `late-fusion', reaches 79.1 cIoU but requires 7B parameters. 
Conversely, SAM with BEIT-3, exemplifying our "early-fusion" approach, achieves a cIoU of 83.7, surpassing all late-fusion methods while utilizing just 1.3 billion parameters.

\subsection{Architecture}

Fig.~\ref{fig:architecture} illustrates the overview of \name{}, which is a simple yet effective framework with three modules: Multimodal Encoder, Projector, and Segment Anything Model~(SAM).

\mypara{Multimodal Encoder.} 
The Early Vision-Language Fused encoder adopts the input image and text and outputs fused multimodal embeddings. In \name{}, we mainly adopt BEIT-3~\citep{beit3} as the Multimodal Encoder, which formulates a multi-way transformer. The text is tokenized by XLMRobertaTokenizer~\citep{xlmRoberta} while the image is resized to \(224^2\) and patched by a 1/16 convolution layer. Within each block of the encoder, the image and text tokens will be fused in the attention block and then fed into separate Feed-Forward Networks~(FFN). We follow ViT~\citep{vit} to retrieve the \texttt{[CLS]} token as the output multimodal embeddings.





\mypara{Adapted prompt encoder for SAM.}
In \name{}, we maintain the architecture of the image encoder and mask decoder of SAM, while extending the prompt encoder to further gather the embeddings from the Multimodal Encoder.
Specifically, the original prompt encoder encodes point or box prompts to \textit{sparse embeddings} of $\mathbb{R}^{B\times N \times D}$, where $B$ is the batch size, $N$ is the number of points/boxes, and $D$ is the embedding dimension. 
In \name{}, the projected multimodal embeddings of $\mathbb{R}^{B \times 1 \times D}$ from the Multimodal Encoder will be concatenated to \textit{sparse embeddings}~(of $\mathbb{R}^{B\times 0 \times D}$, as there is no points/boxes inputs) and then fed into the mask decoder.

\subsection{Data strategy}
\label{data strategy}

Since publicly available open-source RES data is limited, a simple thought is that we use datasets of other segmentation sub-tasks.
To be specific, we include Instance Segmentation~(Objects365~\citep{shao2019objects365}), Semantic Segmentation~(ADE20K~\citep{ade1, ade2}) and Part Segmentation~(Pascal-Part~\citep{pascal_part}, HumanParsing~\citep{liang2015human}, PartImageNet~\citep{he2022partimagenet}) to boost the performance of our \name{}, as well as to extend the model's capability granularity.

However, segmentation data from different tasks exhibit varying distributions, making it challenging to effectively integrate them for joint training.
We carefully analyze the semantic conflict between semantic segmentation data and RES data, and the ambiguity problem between instance segmentation data and RES data in Sec.~\ref{seg:data_challenge}. 
Furthermore, we detailedly explain our data strategies in Sec.~\ref{seg:data_strategy}, including applying a special token `[semantic]' for solving semantic conflict, and using filtering/merging principles to solve the ambiguity problem.

Powered by our data strategies, we successfully integrate segmentation data from diverse tasks into an unified hybrid dataset. 
Upon joint training with the hybrid dataset, we observe the performance gain of \name{} from average 78.0 cIoU to 79.0 cIoU, as shown in Tab.~\ref{tab:benchmark}. 
Furthermore, \name{} can handle various granularities of text-prompted segmentation such as semantic-level~(\eg, stuff, multi-object) and part-level~(\eg, body-part) segmentation.
Detailed definition, metrics and visualizations are shown in Sec.~\ref{sec:semanticRES} and Sec.~\ref{sec:partRES}, separately.

\begin{table}[t]
    \centering
    \small
    \caption{Comparison of the model efficiency of \name{} with LLM-based RES methods. \textbf{Training time} means the total time cost of training on RefCOCO~\citep{referit,refcoco/+}. \textbf{Evaluation time} means the average time cost of forward during evaluation on refcoco/unc/testA.}
    \vspace{-7pt}
    \scalebox{0.85}{
    \begin{tabular}{llll}\toprule
        Methods & Training time & Evaluation time & cIoU\\\midrule
        LISA-7B-LORA & 54.7h & 0.772s & 76.5 \\
        EVF-SAM & 19.7h~\resizebox{!}{0.7\height}{\textcolor{red}{(-64\%)}} & 0.283s~\resizebox{!}{0.7\height}{\textcolor{red}{(-63\%)}} & 84.2~\resizebox{!}{0.7\height}{\textcolor{red}{(+10\%)}} \\
        Effi-EVF-SAM-S & 8.2h~\resizebox{!}{0.7\height}{\textcolor{red}{(-85\%)}} & 0.129s~\resizebox{!}{0.7\height}{\textcolor{red}{(-83\%)}} & 83.5~\resizebox{!}{0.7\height}{\textcolor{red}{(+9\%)}} \\\bottomrule
    \end{tabular}
    }
    \vspace{-15pt}
    \label{tab:efficiency}
\end{table}

\section{Experiments}
\subsection{Datasets and Metrics}
\mypara{Datasets.}
We mainly conduct the experiments on RefCLEF~\citep{referit}, RefCOCO, RefCOCO+~\citep{refcoco/+, referit}, and RefCOCOg~\citep{refcocog, refcocog2}.
Specifically, RefCOCO+ excludes geometric expression~(\eg, `\textit{on the left}'), while RefCOCOg contains longer expressions~(8 words per prompt on average). Among different splits of testing datasets, `testA' is human-centric, while `testB' aims for common objects. 

To produce our best result, we employ multi-task unified training, as illustrated in Sec.~\ref{data strategy}. Objects365~\citep{shao2019objects365}, ADE20K~\citep{ade1, ade2}, Pascal-Part~\citep{pascal_part}, HumanParsing~\citep{liang2015human} and PartImageNet~\citep{he2022partimagenet} are additionally included.

\mypara{Metrics.}
The gIoU and the cIoU are the most commonly calculated metrics on referring expression segmentation benchmarks. The gIoU is the average intersection-over-unions~(IoU) among all images in the test datasets, while the cIoU is the cumulative intersection over the cumulative union.
If not specifically declared, we follow previous works and report the cIoU as the main metric.

\subsection{Implementation Details}
Unless specified, we initialize the proposed \name{} with the public weights of SAM-ViT-Huge~\footnote{SAM: \url{https://github.com/facebookresearch/segment-anything}}~\citep{sam} and BEIT-3-Large~\footnote{BEIT-3: \url{https://github.com/microsoft/unilm}}~\citep{beit3}.
All models are trained on 4 NVIDIA L40s GPUs with mixed precision.
We adopt DeepSpeed~\citep{deepspeed4science} with ZeRO-2 for model parallel to optimize memory consumption.
During training, the batch size of each GPU is 16 and we use gradient accumulation for 2 steps, therefore, the total batch size per iteration is 128.
We adopt AdamW~\citep{adamw} optimizer and set the initial learning rate to 1e-4 with a linear-decay schedule.
We train all models for 15k iterations and use the binary cross-entropy loss (BCE) and dice loss (the weight of both losses is 1.0).

\begin{table*}
    \centering
    \setlength{\abovecaptionskip}{0.2cm}
    \small
    \caption{\textbf{Ablation on fusion methods.} We evaluate the performance of using different pre-trained Multimodal Encoders in \name{}, \eg, CLIP from OpenAI~\citep{clip} or OpenCLIP~\citep{ilharco_gabriel_2021_5143773}. $L_{i}$ denotes the $i$-th layer in the BEIT-3 model~(totally 24 layers for BEIT-3-Large). Half of the layers are activated to assess the impact of the modality fusion stage on model performance. 
    $\dagger$: pre-trained models provided by OpenAI. $\ddagger$: pre-trained models provided by OpenCLIP.}
    \setlength{\tabcolsep}{6pt}
    \scalebox{0.8}{
    \begin{tabular}{lccclccccccccc}\toprule
        \multirow{2}{*}{Encoder} & \multirow{2}{*}{Params}  & \multirow{2}{*}{Text} & \multirow{2}{*}{Image} & \multirow{2}{*}{Modality Fusion} & \multicolumn{3}{c}{RefCOCO} & \multicolumn{3}{c}{RefCOCO+} & \multicolumn{2}{c}{RefCOCOg} & \multirow{2}{*}{AVG}\\\cmidrule{6-13}
        &&&&&val&testA&testB&val&testA&testB&val&test\\
        \midrule
        \multicolumn{6}{l}{\textit{CLIP variants.}} \\
        CLIP-Large$^\dagger$ & 123M  &\checkmark & & - & 61.0 & 63.4 & 59.9 & 43.1 & 45.9 & 40.6 & 48.9 & 49.6 & 51.6\\
        CLIP-Large$^\dagger$ & 428M  &\checkmark & \checkmark & Late (Concat) & 67.4 & 68.9 & 64.4 & 50.5 & 54.6 & 46.7 & 55.1 & 56.2 & 58.0 \\
        CLIP-Large$^\ddagger$ & 123M  &\checkmark & & - & 60.8 & 63.2 & 59.0 & 42.9 & 46.4 & 39.2 & 49.2 & 50.5 & 51.4 \\
        CLIP-Large$^\ddagger$ & 428M  &\checkmark & \checkmark & Late (Concat) & 66.1 & 67.8 & 63.1 & 49.8 & 51.9 & 44.1 & 54.1 & 55.0 & 56.5\\
        CLIP-Huge$^\ddagger$ & 302M  & \checkmark & & - & 61.7 & 64.2 & 60.1 & 44.2 & 47.8 & 40.2 & 49.6 & 50.9 & 52.3\\
        CLIP-Huge$^\ddagger$ & 986M  & \checkmark  & \checkmark & Late (Concat) & 66.3 & 68.2 & 64.3 & 49.8 & 53.5 & 45.1 & 55.4 & 56.7 & 57.4\\\midrule
        \multicolumn{6}{l}{\textit{Early-fused vision-language models.}} \\
        ViLT & 133M& \checkmark & & - & 61.0 & 63.0 & 60.0 & 42.5 & 45.4 & 39.5 & 49.3 & 49.5 & 51.3\\
        ViLT & 136M& \checkmark & \checkmark & Late (Concat) & 61.4 & 64.0 & 59.6 & 42.8 & 46.4 & 40.1 & 49.5 & 50.0 & 51.7\\
        ViLT & 136M & \checkmark & \checkmark & Early & 73.9 & 75.3 & 70.9 & 61.1 & 64.4 & 55.2 & 65.1 & 66.8 & 66.6 \\
        BEIT-3-Large & 370M  & \checkmark & & - & 61.6 & 65.1 & 59.4 & 44.0 & 47.6 & 40.6 & 49.5 & 50.8 & 52.3 \\
        BEIT-3-Large & 673M  &\checkmark & \checkmark & Late (Concat) & 67.7 & 70.2 & 65.4 & 51.1 & 55.0 & 46.9 & 57.2 & 57.0 & 58.8 \\
        BEIT-3-Large & 673M  & \checkmark & \checkmark & Early~($L_1\sim L_{12}$) & 80.6 & 82.2 & 78.8 & 72.4 & 75.7 & 66.7 & 73.7 & 75.0 & 75.6\\
        BEIT-3-Large & 673M  &\checkmark & \checkmark & Early~($L_1\sim L_{24}$) & 82.1 & 83.7 & 80.0 & 75.2 & 78.3 & 70.1 & 76.8 & 77.4 & 78.0\\
        \bottomrule
    \end{tabular}
    }
    \vspace{-15pt}
    \label{tab:fusion}
\end{table*}

\subsection{Main Results}
\label{benchmark}
We mainly report the cIoU metric of RefCOCO/+/g benchmarks and compare our proposed \name{} with recent state-of-the-art methods in Tab.~\ref{tab:benchmark}.
The upper part of Tab.~\ref{tab:benchmark} presents traditional methods based on text encoders. Despite their advantages in terms of fewer parameters or task-specific architectures, these methods either achieve less competitive results or require vast amounts of data and computational resources due to their lack of integration with foundation models. 
The methods listed in the lower portion of Tab.~\ref{tab:benchmark} are based on Large Language Models~(LLMs), achieving more competitive performance but at the expense of significantly increased parameter size. 
Our EVF-SAM achieves the highest average cIoU score across all RES benchmarks, using only readily available data, manageable computation resources, and moderate parameter size. 
Notably, unlike previous methods that excelled on specific datasets but performed poorly on others, our approach achieves consistently high scores across multiple datasets. This demonstrates that our method does not rely on specific linguistic expressions~(\eg, geometric words) for prediction but rather possesses strong generalization capabilities. 


Besides metrics on RefCOCO/+/g, we also evaluate the model's performance on semantic-level referring expression segmentation, part-level referring expression segmentation and reasoning segmentation. Metrics are shown in Sec.~\ref{sec:semanticRES}, Sec.~\ref{sec:partRES} and Sec.~\ref{sec:reasonSeg}.

\subsection{Efficiency analysis}
To prove the efficiency of our \name{} over previous LLM-based models, we present comparisons of training and evaluation time in Tab.~\ref{tab:efficiency}. We take LISA~\citep{lisa}, which employs LLaVA~\citep{llava} as the multimodal extractor, for comparison. 
We follow LISA's reported setting to conduct 10k steps of training with a global batch of 160. 
Considering that different datasets contain descriptions of various lengths~(\eg, 3.6 words per sentence for RefCOCO, 8.4 words per sentence for RefCOCOg), we conduct experiments only on RefCOCO to avoid sample bias. 
Our \name{} can reduce 64\% to 85\% training time cost compared to LLM-based models~(\ie, LISA).
We also record the evaluation time on RefCOCO/testA~(750 images with 5556 referring instances) per image. 
Our \name{} can reduce 63\% to 83\% inference time cost, too.

\subsection{Ablation Study}
In this section, we conduct experiments to investigate the vision-language models for text-prompted SAM and study the effects of the designs of the proposed \name{}. A few ablations are listed in the supplementary. 
Unless specified, we mainly report the cIoU on testA of RefCOCO.

\begin{table*}[t]
    \centering
    \small
    \newcommand{\tg}[1]{\textcolor{gray}{#1}}
    \setlength{\tabcolsep}{6pt}
    \setlength{\abovecaptionskip}{0.2cm}
    \caption{\textbf{Comparison of effects of different foundation models.} The main bottleneck of \name{} lies in the the Multimodal Encoder. Our proposed architecture can perform well on various SAM-like methods.}
    \vspace{-5pt}
    \scalebox{0.85}{
    \begin{tabular}{llcccccccccc}\toprule
        \multirow{2}{*}{Multimodal Encoder} & \multirow{2}{*}{SAM} & \multirow{2}{*}{Params} &  \multicolumn{3}{c}{RefCOCO} & \multicolumn{3}{c}{RefCOCO+} & \multicolumn{2}{c}{RefCOCOg} & \multirow{2}{*}{AVG} \\\cmidrule{4-11}
        &&& val & testA & testB & val & testA & testB & val & test\\
        \midrule
        \tg{CLIP-Large} & \tg{SAM-ViT-H} & \tg{1.08B} & \tg{61.0} & \tg{63.4} & \tg{59.9} & \tg{43.1} & \tg{45.9} & \tg{40.6} & \tg{48.9} & \tg{49.6} & \tg{51.6} \\
        ViLT & SAM-ViT-H &  783M & 73.9 & 75.3 & 70.9 & 61.1 &64.4&55.2&65.1&66.8&66.6\\
        BEIT-3-Base & SAM-ViT-H & 863M & 78.9 & 80.6 & 75.3 & 69.8 & 74.2 & 63.0 & 71.6 & 72.9 & 73.3\\
        BEIT-3-Large & Efficient-SAM-S & 700M & 82.5 & 83.5 & \textbf{80.4} & 75.4 & 77.9 & 70.2 & 76.1 & 77.1 & 77.9\\
        BEIT-3-Large & SAM-ViT-H & 1.32B & 82.1 & 83.7 & 80.0 & 75.2 & 78.3 & 70.1 & 76.8 & 77.4 & 78.0 \\
        BEIT-3-Large & SAM-2-L & 898M & \textbf{82.7} & \textbf{84.1} & 80.0 & \textbf{76.3} & \textbf{80.1} & \textbf{71.8} & \textbf{77.0
} & \textbf{78.4} & \textbf{78.8} \\
        \bottomrule 
    \end{tabular}
    }
    \vspace{-15pt}
    \label{tab:lightweight}
\end{table*}

\mypara{Multimodal Encoder and fusion methods.} 
\label{fusion stage}
In Tab.~\ref{tab:fusion}, we explore the effects of different Multimodal Encoders, \eg, CLIP, ViLT~\citep{kim2021vilt}, and BEIT-3, and fusion methods, \eg, late fusion or early fusion.
As shown in Tab.~\ref{tab:fusion}, using a text-only encoder in \name{} obtains limited segmentation performance on RefCOCO.
Using Multimodal Encoders with both image and text inputs remarkably improves 4.5 cIoU, 4.6 cIoU, 4.0 cIoU, 1.0 cIoU, and 4.5 cIoU for CLIP-Large$^\dagger$ (OpenAI~\footnote{OpenAI: \url{https://github.com/openai/CLIP}}), CLIP-Large$^\ddagger$ (OpenCLIP~\footnote{OpenCLIP: \url{https://github.com/mlfoundations/open_clip}}), CLIP-Huge$^\ddagger$ (OpenCLIP), ViLT, and BEIT-3, respectively.
It demonstrates the superiority of using multimodal prompts (text and input image) and showcases that the image embeddings will also provide useful guidance for SAM to segment objects accurately.
We further evaluate the effects of early fusion on ViLT and BEIT-3, which adopts modality fusions in all self-attention layers.
Specifically, we adopt two settings for BEIT-3 to analyze, \eg, fusions among former 12 layers ($L_1\sim L_{12}$), and fusions among all layers ($L_1\sim L_{24}$).
Tab.~\ref{tab:fusion} indicates that BEIT-3 with early fusion (fusing former 12 layers or fusing all 24 layers) significantly improves compared to late fusion or using text only.
In addition, ViLT with early fusion also achieves 11.1 cIoU improvements compared to the baseline with text-only prompts, showing the effectiveness of early fusion and multimodal inputs for prompting SAM.
Therefore, Tab.~\ref{tab:fusion} demonstrates that \textit{(1) Multimodal Encoder with the input image and text and (2) early fusions between the image and text encoder} are much effective for text-prompted SAM.

\begin{table}[t]
\centering
\small
\caption{\textbf{Ablation on trainable modules.}
We mainly evaluate the effects of fine-tuning or freezing the Multimodal Encoder, the prompt encoder and mask decoder of SAM. `\checkmark' denotes trainable, while `$\ast$' denotes frozen.
\vspace{-10pt}
}
\setlength{\tabcolsep}{8pt}
\scalebox{0.85}{
\begin{tabular}{cccc}\toprule
    Multimodal Enc. & Prompt Enc. & Mask Dec. & \makecell{cIoU} \\
    \midrule
    $\ast$ &  $\ast$ & \checkmark & 21.2\\
    \checkmark & $\ast$ & $\ast$ & 82.9\\
    \checkmark & $\ast$ & \checkmark & 83.3\\
    \checkmark & \checkmark & \checkmark & \textbf{83.7}\\
    \bottomrule
\end{tabular}}
\vspace{-18pt}
\label{tab:trainable}
\end{table}

\mypara{Ablation on trainable modules.} 
\label{trainable}
In Tab.~\ref{tab:trainable}, we evaluated the effects of fine-tuning (\checkmark) or freezing ($\ast$) modules in the proposed \name{}, \ie, the Multimodal Encoder, the prompt encoder, and the mask decoder.
The image encoder of SAM is kept frozen during training.
As Tab.~\ref{tab:trainable} shows, fine-tuning the Multimodal Encoder is crucial and it adapts the Multimodal Encoder to encode text and image inputs to multimodal representation for referring image segmentation.
Notably, \name{} can achieve competitive results with all modules of SAM kept frozen, and it can be seamlessly regarded as a strong extension for the original SAM, which simultaneously supports text prompts, box prompts and point prompts.
Tab.~\ref{tab:trainable} shows fine-tuning the prompt encoder and mask decoder of SAM brings significant improvements.


\mypara{Effects of Different Foundation Models.}
In Tab.~\ref{tab:lightweight}, we explore the effects of using different foundation models in \name{}.
For the Multimodal Encoder, we adopt CLIP-Large (only text encoder), ViLT, BEIT-3-Large, and BEIT-3-Base.
We also modify \name{} with Efficient-SAM~\citep{efficientsam} and SAM-2~\citep{ravi2024sam2} to formulate a lighter or better version.
As shown in Tab.~\ref{tab:lightweight}, \name{} with BEIT-3-Base brings a severe performance drop which indicates a better Multimodal Encoder leads to better prompts for SAM.
Remarkably, Tab.~\ref{tab:lightweight} shows a negligible difference between Efficient-SAM-S and SAM-H in \name{}, which indicates that \name{} performs well for different SAM variants.
In addition, it also provides insights about designing text-prompted SAMs for future research, \eg, \textit{developing a larger and better Multimodal Encoder is more important to empower SAM with text-prompted abilities}.


\section{Conclusion}
In this paper, we explore the effective ways to prompt SAM with texts and demonstrate the importance of using the Multimodal Encoder with early-fusion and multimodal inputs.
To this end, we propose \name{}, which establishes a new and simple path for extending SAMs' text-prompted segmentation abilities with the \textit{off-the-shelf} foundation models.
We conduct experiments on the Referring Expression Segmentation~(RES) tasks with various benchmarks to evaluate the performance of text-prompted SAM.
Experimental results showcase that our \name{} achieves state-of-the-art performance for segmenting objects with referring texts on RefCOCO/+/g benchmarks, outperforming recent approaches based on Large Language Models with huge numbers of parameters.
Moreover, we propose carefully-designed data strategies for balancing conflicts from data of different distributions. We construct a harmonious hybrid dataset, leading to an uni-model that performs well on various segmentation tasks.
We hope this study can bring new ideas or insights to inspire future research on prompting SAM with texts.




{
    \small
    \bibliographystyle{ieeenat_fullname}
    \bibliography{main}

\begin{thebibliography}{79}
\providecommand{\natexlab}[1]{#1}
\providecommand{\url}[1]{\texttt{#1}}
\expandafter\ifx\csname urlstyle\endcsname\relax
  \providecommand{\doi}[1]{doi: #1}\else
  \providecommand{\doi}{doi: \begingroup \urlstyle{rm}\Url}\fi

\bibitem[Bai et~al.(2023)Bai, Bai, Yang, Wang, Tan, Wang, Lin, Zhou, and Zhou]{Qwen-VL}
Jinze Bai, Shuai Bai, Shusheng Yang, Shijie Wang, Sinan Tan, Peng Wang, Junyang Lin, Chang Zhou, and Jingren Zhou.
\newblock Qwen-vl: A versatile vision-language model for understanding, localization, text reading, and beyond.
\newblock \emph{arXiv preprint arXiv:2308.12966}, 2023.

\bibitem[Bavishi et~al.(2023)Bavishi, Elsen, Hawthorne, Nye, Odena, Somani, and Ta\c{s}\i{}rlar]{fuyu-8b}
Rohan Bavishi, Erich Elsen, Curtis Hawthorne, Maxwell Nye, Augustus Odena, Arushi Somani, and Sa\u{g}nak Ta\c{s}\i{}rlar.
\newblock Introducing our multimodal models, 2023.

\bibitem[Caesar et~al.(2018)Caesar, Uijlings, and Ferrari]{cocostuff}
Holger Caesar, Jasper Uijlings, and Vittorio Ferrari.
\newblock Coco-stuff: Thing and stuff classes in context.
\newblock In \emph{Proceedings of the IEEE conference on computer vision and pattern recognition}, pages 1209--1218, 2018.

\bibitem[Chen et~al.(2019)Chen, Jia, Lo, Chen, and Liu]{chen2019see}
Ding-Jie Chen, Songhao Jia, Yi-Chen Lo, Hwann-Tzong Chen, and Tyng-Luh Liu.
\newblock See-through-text grouping for referring image segmentation.
\newblock In \emph{Proceedings of the IEEE/CVF International Conference on Computer Vision}, pages 7454--7463, 2019.

\bibitem[Chen et~al.(2014)Chen, Mottaghi, Liu, Fidler, Urtasun, and Yuille]{pascal_part}
Xianjie Chen, Roozbeh Mottaghi, Xiaobai Liu, Sanja Fidler, Raquel Urtasun, and Alan Yuille.
\newblock Detect what you can: Detecting and representing objects using holistic models and body parts.
\newblock In \emph{Proceedings of the IEEE conference on computer vision and pattern recognition}, pages 1971--1978, 2014.

\bibitem[Conneau et~al.(2019)Conneau, Khandelwal, Goyal, Chaudhary, Wenzek, Guzm{\'a}n, Grave, Ott, Zettlemoyer, and Stoyanov]{xlmRoberta}
Alexis Conneau, Kartikay Khandelwal, Naman Goyal, Vishrav Chaudhary, Guillaume Wenzek, Francisco Guzm{\'a}n, Edouard Grave, Myle Ott, Luke Zettlemoyer, and Veselin Stoyanov.
\newblock Unsupervised cross-lingual representation learning at scale.
\newblock \emph{arXiv preprint arXiv:1911.02116}, 2019.

\bibitem[Devlin et~al.(2018)Devlin, Chang, Lee, and Toutanova]{bert}
Jacob Devlin, Ming-Wei Chang, Kenton Lee, and Kristina Toutanova.
\newblock Bert: Pre-training of deep bidirectional transformers for language understanding.
\newblock \emph{arXiv preprint arXiv:1810.04805}, 2018.

\bibitem[Diao et~al.(2025)Diao, Cui, Li, Wang, Lu, and Wang]{EVE}
Haiwen Diao, Yufeng Cui, Xiaotong Li, Yueze Wang, Huchuan Lu, and Xinlong Wang.
\newblock Unveiling encoder-free vision-language models.
\newblock \emph{Advances in Neural Information Processing Systems}, 37:\penalty0 52545--52567, 2025.

\bibitem[Ding et~al.(2021)Ding, Liu, Wang, and Jiang]{ding2021vision}
Henghui Ding, Chang Liu, Suchen Wang, and Xudong Jiang.
\newblock Vision-language transformer and query generation for referring segmentation.
\newblock In \emph{Proceedings of the IEEE/CVF International Conference on Computer Vision}, pages 16321--16330, 2021.

\bibitem[Dosovitskiy et~al.(2020)Dosovitskiy, Beyer, Kolesnikov, Weissenborn, Zhai, Unterthiner, Dehghani, Minderer, Heigold, Gelly, et~al.]{vit}
Alexey Dosovitskiy, Lucas Beyer, Alexander Kolesnikov, Dirk Weissenborn, Xiaohua Zhai, Thomas Unterthiner, Mostafa Dehghani, Matthias Minderer, Georg Heigold, Sylvain Gelly, et~al.
\newblock An image is worth 16x16 words: Transformers for image recognition at scale.
\newblock \emph{arXiv preprint arXiv:2010.11929}, 2020.

\bibitem[Everingham et~al.(2010)Everingham, Van~Gool, Williams, Winn, and Zisserman]{everingham2010pascal}
Mark Everingham, Luc Van~Gool, Christopher~KI Williams, John Winn, and Andrew Zisserman.
\newblock The pascal visual object classes (voc) challenge.
\newblock \emph{International journal of computer vision}, 88:\penalty0 303--338, 2010.

\bibitem[He et~al.(2022)He, Yang, Yang, Kortylewski, Yuan, Chen, Liu, Yang, Yu, and Yuille]{he2022partimagenet}
Ju He, Shuo Yang, Shaokang Yang, Adam Kortylewski, Xiaoding Yuan, Jie-Neng Chen, Shuai Liu, Cheng Yang, Qihang Yu, and Alan Yuille.
\newblock Partimagenet: A large, high-quality dataset of parts.
\newblock In \emph{European Conference on Computer Vision}, pages 128--145. Springer, 2022.

\bibitem[Hu et~al.(2016)Hu, Rohrbach, and Darrell]{hu2016segmentation}
Ronghang Hu, Marcus Rohrbach, and Trevor Darrell.
\newblock Segmentation from natural language expressions.
\newblock In \emph{Computer Vision--ECCV 2016: 14th European Conference, Amsterdam, The Netherlands, October 11--14, 2016, Proceedings, Part I 14}, pages 108--124. Springer, 2016.

\bibitem[Hu et~al.(2020)Hu, Feng, Sun, Zhang, and Lu]{hu2020bi}
Zhiwei Hu, Guang Feng, Jiayu Sun, Lihe Zhang, and Huchuan Lu.
\newblock Bi-directional relationship inferring network for referring image segmentation.
\newblock In \emph{Proceedings of the IEEE/CVF conference on computer vision and pattern recognition}, pages 4424--4433, 2020.

\bibitem[Huang et~al.(2020)Huang, Zeng, Liu, Fu, and Fu]{huang2020pixel-bert}
Zhicheng Huang, Zhaoyang Zeng, Bei Liu, Dongmei Fu, and Jianlong Fu.
\newblock Pixel-bert: Aligning image pixels with text by deep multi-modal transformers.
\newblock \emph{arXiv preprint arXiv:2004.00849}, 2020.

\bibitem[Hudson and Manning(2019)]{hudson2019gqa}
Drew~A Hudson and Christopher~D Manning.
\newblock Gqa: A new dataset for real-world visual reasoning and compositional question answering.
\newblock In \emph{Proceedings of the IEEE/CVF conference on computer vision and pattern recognition}, pages 6700--6709, 2019.

\bibitem[Ilharco et~al.(2021)Ilharco, Wortsman, Wightman, Gordon, Carlini, Taori, Dave, Shankar, Namkoong, Miller, Hajishirzi, Farhadi, and Schmidt]{ilharco_gabriel_2021_5143773}
Gabriel Ilharco, Mitchell Wortsman, Ross Wightman, Cade Gordon, Nicholas Carlini, Rohan Taori, Achal Dave, Vaishaal Shankar, Hongseok Namkoong, John Miller, Hannaneh Hajishirzi, Ali Farhadi, and Ludwig Schmidt.
\newblock Openclip, 2021.
\newblock If you use this software, please cite it as below.

\bibitem[Jocher et~al.(2023)Jocher, Chaurasia, and Qiu]{yolov8}
Glenn Jocher, Ayush Chaurasia, and Jing Qiu.
\newblock {Ultralytics YOLO}, 2023.

\bibitem[Johnson et~al.(2017)Johnson, Hariharan, Van Der~Maaten, Fei-Fei, Lawrence~Zitnick, and Girshick]{johnson2017clevr}
Justin Johnson, Bharath Hariharan, Laurens Van Der~Maaten, Li Fei-Fei, C Lawrence~Zitnick, and Ross Girshick.
\newblock Clevr: A diagnostic dataset for compositional language and elementary visual reasoning.
\newblock In \emph{Proceedings of the IEEE conference on computer vision and pattern recognition}, pages 2901--2910, 2017.

\bibitem[Kamath et~al.(2021)Kamath, Singh, LeCun, Synnaeve, Misra, and Carion]{GoldG}
Aishwarya Kamath, Mannat Singh, Yann LeCun, Gabriel Synnaeve, Ishan Misra, and Nicolas Carion.
\newblock Mdetr-modulated detection for end-to-end multi-modal understanding.
\newblock In \emph{Proceedings of the IEEE/CVF international conference on computer vision}, pages 1780--1790, 2021.

\bibitem[Kazemzadeh et~al.(2014)Kazemzadeh, Ordonez, Matten, and Berg]{referit}
Sahar Kazemzadeh, Vicente Ordonez, Mark Matten, and Tamara Berg.
\newblock Referitgame: Referring to objects in photographs of natural scenes.
\newblock In \emph{Proceedings of the 2014 conference on empirical methods in natural language processing (EMNLP)}, pages 787--798, 2014.

\bibitem[Ke et~al.(2024)Ke, Ye, Danelljan, Tai, Tang, Yu, et~al.]{samhq}
Lei Ke, Mingqiao Ye, Martin Danelljan, Yu-Wing Tai, Chi-Keung Tang, Fisher Yu, et~al.
\newblock Segment anything in high quality.
\newblock \emph{Advances in Neural Information Processing Systems}, 36, 2024.

\bibitem[Kim et~al.(2021)Kim, Son, and Kim]{kim2021vilt}
Wonjae Kim, Bokyung Son, and Ildoo Kim.
\newblock Vilt: Vision-and-language transformer without convolution or region supervision.
\newblock In \emph{International conference on machine learning}, pages 5583--5594. PMLR, 2021.

\bibitem[Kirillov et~al.(2023)Kirillov, Mintun, Ravi, Mao, Rolland, Gustafson, Xiao, Whitehead, Berg, Lo, et~al.]{sam}
Alexander Kirillov, Eric Mintun, Nikhila Ravi, Hanzi Mao, Chloe Rolland, Laura Gustafson, Tete Xiao, Spencer Whitehead, Alexander~C Berg, Wan-Yen Lo, et~al.
\newblock Segment anything.
\newblock In \emph{Proceedings of the IEEE/CVF International Conference on Computer Vision}, pages 4015--4026, 2023.

\bibitem[Lai et~al.(2023)Lai, Tian, Chen, Li, Yuan, Liu, and Jia]{lisa}
Xin Lai, Zhuotao Tian, Yukang Chen, Yanwei Li, Yuhui Yuan, Shu Liu, and Jiaya Jia.
\newblock Lisa: Reasoning segmentation via large language model.
\newblock \emph{arXiv preprint arXiv:2308.00692}, 2023.

\bibitem[Li and Sigal(2021{\natexlab{a}})]{li2021referring}
Muchen Li and Leonid Sigal.
\newblock Referring transformer: A one-step approach to multi-task visual grounding.
\newblock \emph{Advances in neural information processing systems}, 34:\penalty0 19652--19664, 2021{\natexlab{a}}.

\bibitem[Li and Sigal(2021{\natexlab{b}})]{li2021reftr}
Muchen Li and Leonid Sigal.
\newblock Referring transformer: A one-step approach to multi-task visual grounding.
\newblock \emph{Advances in neural information processing systems}, 34:\penalty0 19652--19664, 2021{\natexlab{b}}.

\bibitem[Li et~al.(2020)Li, Wei, Chen, Tai, and Tang]{li2020fss}
Xiang Li, Tianhan Wei, Yau~Pun Chen, Yu-Wing Tai, and Chi-Keung Tang.
\newblock Fss-1000: A 1000-class dataset for few-shot segmentation.
\newblock In \emph{Proceedings of the IEEE/CVF conference on computer vision and pattern recognition}, pages 2869--2878, 2020.

\bibitem[Li et~al.(2023{\natexlab{a}})Li, Bubeck, Eldan, Del~Giorno, Gunasekar, and Lee]{phi15}
Yuanzhi Li, S{\'e}bastien Bubeck, Ronen Eldan, Allie Del~Giorno, Suriya Gunasekar, and Yin~Tat Lee.
\newblock Textbooks are all you need ii: phi-1.5 technical report.
\newblock \emph{arXiv preprint arXiv:2309.05463}, 2023{\natexlab{a}}.

\bibitem[Li et~al.(2023{\natexlab{b}})Li, Zhang, Teng, and Lan]{refsam}
Yonglin Li, Jing Zhang, Xiao Teng, and Long Lan.
\newblock Refsam: Efficiently adapting segmenting anything model for referring video object segmentation.
\newblock \emph{arXiv preprint arXiv:2307.00997}, 2023{\natexlab{b}}.

\bibitem[Liang et~al.(2015{\natexlab{a}})Liang, Liu, Shen, Yang, Liu, Dong, Lin, and Yan]{liang2015deep}
Xiaodan Liang, Si Liu, Xiaohui Shen, Jianchao Yang, Luoqi Liu, Jian Dong, Liang Lin, and Shuicheng Yan.
\newblock Deep human parsing with active template regression.
\newblock \emph{IEEE transactions on pattern analysis and machine intelligence}, 37\penalty0 (12):\penalty0 2402--2414, 2015{\natexlab{a}}.

\bibitem[Liang et~al.(2015{\natexlab{b}})Liang, Xu, Shen, Yang, Liu, Tang, Lin, and Yan]{liang2015human}
Xiaodan Liang, Chunyan Xu, Xiaohui Shen, Jianchao Yang, Si Liu, Jinhui Tang, Liang Lin, and Shuicheng Yan.
\newblock Human parsing with contextualized convolutional neural network.
\newblock In \emph{Proceedings of the IEEE international conference on computer vision}, pages 1386--1394, 2015{\natexlab{b}}.

\bibitem[Lin et~al.(2014)Lin, Maire, Belongie, Hays, Perona, Ramanan, Doll{\'a}r, and Zitnick]{coco}
Tsung-Yi Lin, Michael Maire, Serge Belongie, James Hays, Pietro Perona, Deva Ramanan, Piotr Doll{\'a}r, and C~Lawrence Zitnick.
\newblock Microsoft coco: Common objects in context.
\newblock In \emph{Computer Vision--ECCV 2014: 13th European Conference, Zurich, Switzerland, September 6-12, 2014, Proceedings, Part V 13}, pages 740--755. Springer, 2014.

\bibitem[Liu et~al.(2017)Liu, Lin, Shen, Yang, Lu, and Yuille]{liu2017recurrent}
Chenxi Liu, Zhe Lin, Xiaohui Shen, Jimei Yang, Xin Lu, and Alan Yuille.
\newblock Recurrent multimodal interaction for referring image segmentation.
\newblock In \emph{Proceedings of the IEEE international conference on computer vision}, pages 1271--1280, 2017.

\bibitem[Liu et~al.(2023{\natexlab{a}})Liu, Ding, and Jiang]{liu2023gres}
Chang Liu, Henghui Ding, and Xudong Jiang.
\newblock Gres: Generalized referring expression segmentation.
\newblock In \emph{Proceedings of the IEEE/CVF conference on computer vision and pattern recognition}, pages 23592--23601, 2023{\natexlab{a}}.

\bibitem[Liu et~al.(2023{\natexlab{b}})Liu, Li, Wu, and Lee]{llava}
Haotian Liu, Chunyuan Li, Qingyang Wu, and Yong~Jae Lee.
\newblock Visual instruction tuning, 2023{\natexlab{b}}.

\bibitem[Liu et~al.(2024)Liu, Li, Li, and Lee]{llava1.5}
Haotian Liu, Chunyuan Li, Yuheng Li, and Yong~Jae Lee.
\newblock Improved baselines with visual instruction tuning.
\newblock In \emph{Proceedings of the IEEE/CVF Conference on Computer Vision and Pattern Recognition}, pages 26296--26306, 2024.

\bibitem[Liu et~al.(2023{\natexlab{c}})Liu, Ding, Cai, Zhang, Satzoda, Mahadevan, and Manmatha]{polyformer}
Jiang Liu, Hui Ding, Zhaowei Cai, Yuting Zhang, Ravi~Kumar Satzoda, Vijay Mahadevan, and R Manmatha.
\newblock Polyformer: Referring image segmentation as sequential polygon generation.
\newblock In \emph{Proceedings of the IEEE/CVF Conference on Computer Vision and Pattern Recognition}, pages 18653--18663, 2023{\natexlab{c}}.

\bibitem[Liu et~al.(2023{\natexlab{d}})Liu, Zhang, Wang, Wang, Yang, and Tang]{uniseg}
Yong Liu, Cairong Zhang, Yitong Wang, Jiahao Wang, Yujiu Yang, and Yansong Tang.
\newblock Universal segmentation at arbitrary granularity with language instruction.
\newblock \emph{arXiv preprint arXiv:2312.01623}, 2023{\natexlab{d}}.

\bibitem[Loshchilov and Hutter(2017)]{adamw}
Ilya Loshchilov and Frank Hutter.
\newblock Decoupled weight decay regularization.
\newblock \emph{arXiv preprint arXiv:1711.05101}, 2017.

\bibitem[Mao et~al.(2016)Mao, Huang, Toshev, Camburu, Yuille, and Murphy]{refcocog2}
Junhua Mao, Jonathan Huang, Alexander Toshev, Oana Camburu, Alan~L Yuille, and Kevin Murphy.
\newblock Generation and comprehension of unambiguous object descriptions.
\newblock In \emph{Proceedings of the IEEE conference on computer vision and pattern recognition}, pages 11--20, 2016.

\bibitem[Nagaraja et~al.(2016)Nagaraja, Morariu, and Davis]{refcocog}
Varun~K Nagaraja, Vlad~I Morariu, and Larry~S Davis.
\newblock Modeling context between objects for referring expression understanding.
\newblock In \emph{Computer Vision--ECCV 2016: 14th European Conference, Amsterdam, The Netherlands, October 11--14, 2016, Proceedings, Part IV 14}, pages 792--807. Springer, 2016.

\bibitem[Neuhold et~al.(2017)Neuhold, Ollmann, Rota~Bulo, and Kontschieder]{mapillary}
Gerhard Neuhold, Tobias Ollmann, Samuel Rota~Bulo, and Peter Kontschieder.
\newblock The mapillary vistas dataset for semantic understanding of street scenes.
\newblock In \emph{Proceedings of the IEEE international conference on computer vision}, pages 4990--4999, 2017.

\bibitem[Pi et~al.(2023)Pi, Yao, Gao, Zhang, and Zhang]{perceptiongpt}
Renjie Pi, Lewei Yao, Jiahui Gao, Jipeng Zhang, and Tong Zhang.
\newblock Perceptiongpt: Effectively fusing visual perception into llm.
\newblock \emph{arXiv preprint arXiv:2311.06612}, 2023.

\bibitem[Radford et~al.(2021)Radford, Kim, Hallacy, Ramesh, Goh, Agarwal, Sastry, Askell, Mishkin, Clark, et~al.]{clip}
Alec Radford, Jong~Wook Kim, Chris Hallacy, Aditya Ramesh, Gabriel Goh, Sandhini Agarwal, Girish Sastry, Amanda Askell, Pamela Mishkin, Jack Clark, et~al.
\newblock Learning transferable visual models from natural language supervision.
\newblock In \emph{International conference on machine learning}, pages 8748--8763. PMLR, 2021.

\bibitem[Ramanathan et~al.(2023)Ramanathan, Kalia, Petrovic, Wen, Zheng, Guo, Wang, Marquez, Kovvuri, Kadian, et~al.]{ramanathan2023paco}
Vignesh Ramanathan, Anmol Kalia, Vladan Petrovic, Yi Wen, Baixue Zheng, Baishan Guo, Rui Wang, Aaron Marquez, Rama Kovvuri, Abhishek Kadian, et~al.
\newblock Paco: Parts and attributes of common objects.
\newblock In \emph{Proceedings of the IEEE/CVF Conference on Computer Vision and Pattern Recognition}, pages 7141--7151, 2023.

\bibitem[Rasheed et~al.(2023)Rasheed, Maaz, Shaji, Shaker, Khan, Cholakkal, Anwer, Xing, Yang, and Khan]{glamm}
Hanoona Rasheed, Muhammad Maaz, Sahal Shaji, Abdelrahman Shaker, Salman Khan, Hisham Cholakkal, Rao~M Anwer, Erix Xing, Ming-Hsuan Yang, and Fahad~S Khan.
\newblock Glamm: Pixel grounding large multimodal model.
\newblock \emph{arXiv preprint arXiv:2311.03356}, 2023.

\bibitem[Ravi et~al.(2024)Ravi, Gabeur, Hu, Hu, Ryali, Ma, Khedr, R{\"a}dle, Rolland, Gustafson, Mintun, Pan, Alwala, Carion, Wu, Girshick, Doll{\'a}r, and Feichtenhofer]{ravi2024sam2}
Nikhila Ravi, Valentin Gabeur, Yuan-Ting Hu, Ronghang Hu, Chaitanya Ryali, Tengyu Ma, Haitham Khedr, Roman R{\"a}dle, Chloe Rolland, Laura Gustafson, Eric Mintun, Junting Pan, Kalyan~Vasudev Alwala, Nicolas Carion, Chao-Yuan Wu, Ross Girshick, Piotr Doll{\'a}r, and Christoph Feichtenhofer.
\newblock Sam 2: Segment anything in images and videos.
\newblock \emph{arXiv preprint arXiv:2408.00714}, 2024.

\bibitem[Ren et~al.(2024)Ren, Liu, Zeng, Lin, Li, Cao, Chen, Huang, Chen, Yan, Zeng, Zhang, Li, Yang, Li, Jiang, and Zhang]{groundedSAM}
Tianhe Ren, Shilong Liu, Ailing Zeng, Jing Lin, Kunchang Li, He Cao, Jiayu Chen, Xinyu Huang, Yukang Chen, Feng Yan, Zhaoyang Zeng, Hao Zhang, Feng Li, Jie Yang, Hongyang Li, Qing Jiang, and Lei Zhang.
\newblock Grounded sam: Assembling open-world models for diverse visual tasks, 2024.

\bibitem[Ren et~al.(2023)Ren, Huang, Wei, Zhao, Fu, Feng, and Jin]{pixellm}
Zhongwei Ren, Zhicheng Huang, Yunchao Wei, Yao Zhao, Dongmei Fu, Jiashi Feng, and Xiaojie Jin.
\newblock Pixellm: Pixel reasoning with large multimodal model.
\newblock \emph{arXiv preprint arXiv:2312.02228}, 2023.

\bibitem[Shao et~al.(2019)Shao, Li, Zhang, Peng, Yu, Zhang, Li, and Sun]{shao2019objects365}
Shuai Shao, Zeming Li, Tianyuan Zhang, Chao Peng, Gang Yu, Xiangyu Zhang, Jing Li, and Jian Sun.
\newblock Objects365: A large-scale, high-quality dataset for object detection.
\newblock In \emph{Proceedings of the IEEE/CVF international conference on computer vision}, pages 8430--8439, 2019.

\bibitem[Shi et~al.(2018)Shi, Li, Meng, and Wu]{shi2018key}
Hengcan Shi, Hongliang Li, Fanman Meng, and Qingbo Wu.
\newblock Key-word-aware network for referring expression image segmentation.
\newblock In \emph{Proceedings of the European Conference on Computer Vision (ECCV)}, pages 38--54, 2018.

\bibitem[Song et~al.(2023)Song, Kruft, Zhang, Li, Chen, Zhang, Tanaka, Wu, Rasley, Awan, et~al.]{deepspeed4science}
Shuaiwen~Leon Song, Bonnie Kruft, Minjia Zhang, Conglong Li, Shiyang Chen, Chengming Zhang, Masahiro Tanaka, Xiaoxia Wu, Jeff Rasley, Ammar~Ahmad Awan, et~al.
\newblock Deepspeed4science initiative: Enabling large-scale scientific discovery through sophisticated ai system technologies.
\newblock \emph{arXiv preprint arXiv:2310.04610}, 2023.

\bibitem[Sun et~al.(2023{\natexlab{a}})Sun, Cui, Zhang, Zhang, Yu, Luo, Wang, Rao, Liu, Huang, et~al.]{emu02}
Quan Sun, Yufeng Cui, Xiaosong Zhang, Fan Zhang, Qiying Yu, Zhengxiong Luo, Yueze Wang, Yongming Rao, Jingjing Liu, Tiejun Huang, et~al.
\newblock Generative multimodal models are in-context learners.
\newblock \emph{arXiv preprint arXiv:2312.13286}, 2023{\natexlab{a}}.

\bibitem[Sun et~al.(2023{\natexlab{b}})Sun, Yu, Cui, Zhang, Zhang, Wang, Gao, Liu, Huang, and Wang]{emu01}
Quan Sun, Qiying Yu, Yufeng Cui, Fan Zhang, Xiaosong Zhang, Yueze Wang, Hongcheng Gao, Jingjing Liu, Tiejun Huang, and Xinlong Wang.
\newblock Generative pretraining in multimodality.
\newblock \emph{arXiv preprint arXiv:2307.05222}, 2023{\natexlab{b}}.

\bibitem[Sun et~al.(2024)Sun, Chen, Zhang, Zhang, Chen, Zhang, Ding, Wang, and Li]{sun2024vrp}
Yanpeng Sun, Jiahui Chen, Shan Zhang, Xinyu Zhang, Qiang Chen, Gang Zhang, Errui Ding, Jingdong Wang, and Zechao Li.
\newblock Vrp-sam: Sam with visual reference prompt.
\newblock In \emph{Proceedings of the IEEE/CVF Conference on Computer Vision and Pattern Recognition}, pages 23565--23574, 2024.

\bibitem[Touvron et~al.(2023)Touvron, Lavril, Izacard, Martinet, Lachaux, Lacroix, Rozi{\`e}re, Goyal, Hambro, Azhar, et~al.]{llama}
Hugo Touvron, Thibaut Lavril, Gautier Izacard, Xavier Martinet, Marie-Anne Lachaux, Timoth{\'e}e Lacroix, Baptiste Rozi{\`e}re, Naman Goyal, Eric Hambro, Faisal Azhar, et~al.
\newblock Llama: Open and efficient foundation language models.
\newblock \emph{arXiv preprint arXiv:2302.13971}, 2023.

\bibitem[Wang et~al.(2023)Wang, Wang, Lin, Bai, Zhou, Zhou, Wang, and Zhou]{one-peace}
Peng Wang, Shijie Wang, Junyang Lin, Shuai Bai, Xiaohuan Zhou, Jingren Zhou, Xinggang Wang, and Chang Zhou.
\newblock One-peace: Exploring one general representation model toward unlimited modalities.
\newblock \emph{arXiv preprint arXiv:2305.11172}, 2023.

\bibitem[Wang et~al.(2022{\natexlab{a}})Wang, Bao, Dong, Bjorck, Peng, Liu, Aggarwal, Mohammed, Singhal, Som, et~al.]{beit3}
Wenhui Wang, Hangbo Bao, Li Dong, Johan Bjorck, Zhiliang Peng, Qiang Liu, Kriti Aggarwal, Owais~Khan Mohammed, Saksham Singhal, Subhojit Som, et~al.
\newblock Image as a foreign language: Beit pretraining for all vision and vision-language tasks.
\newblock \emph{arXiv preprint arXiv:2208.10442}, 2022{\natexlab{a}}.

\bibitem[Wang et~al.(2022{\natexlab{b}})Wang, Lu, Li, Tao, Guo, Gong, and Liu]{wang2022cris}
Zhaoqing Wang, Yu Lu, Qiang Li, Xunqiang Tao, Yandong Guo, Mingming Gong, and Tongliang Liu.
\newblock Cris: Clip-driven referring image segmentation.
\newblock In \emph{Proceedings of the IEEE/CVF conference on computer vision and pattern recognition}, pages 11686--11695, 2022{\natexlab{b}}.

\bibitem[Wu et~al.(2023)Wu, Jiang, Yan, Lu, Yuan, and Luo]{uniref++}
Jiannan Wu, Yi Jiang, Bin Yan, Huchuan Lu, Zehuan Yuan, and Ping Luo.
\newblock Uniref++: Segment every reference object in spatial and temporal spaces.
\newblock \emph{arXiv preprint arXiv:2312.15715}, 2023.

\bibitem[Xia et~al.(2023)Xia, Han, Han, Pan, Song, and Huang]{gsva}
Zhuofan Xia, Dongchen Han, Yizeng Han, Xuran Pan, Shiji Song, and Gao Huang.
\newblock Gsva: Generalized segmentation via multimodal large language models.
\newblock \emph{arXiv preprint arXiv:2312.10103}, 2023.

\bibitem[Xie et~al.(2023)Xie, Wang, Ma, Chen, Lu, Yang, Shi, and Lin]{xie2023edit}
Defeng Xie, Ruichen Wang, Jian Ma, Chen Chen, Haonan Lu, Dong Yang, Fobo Shi, and Xiaodong Lin.
\newblock Edit everything: A text-guided generative system for images editing.
\newblock \emph{arXiv preprint arXiv:2304.14006}, 2023.

\bibitem[Xiong et~al.(2023)Xiong, Varadarajan, Wu, Xiang, Xiao, Zhu, Dai, Wang, Sun, Iandola, et~al.]{efficientsam}
Yunyang Xiong, Bala Varadarajan, Lemeng Wu, Xiaoyu Xiang, Fanyi Xiao, Chenchen Zhu, Xiaoliang Dai, Dilin Wang, Fei Sun, Forrest Iandola, et~al.
\newblock Efficientsam: Leveraged masked image pretraining for efficient segment anything.
\newblock \emph{arXiv preprint arXiv:2312.00863}, 2023.

\bibitem[Xu et~al.(2023)Xu, Xu, Yang, Li, Xie, Huang, and Li]{u-llava}
Jinjin Xu, Liwu Xu, Yuzhe Yang, Xiang Li, Yanchun Xie, Yi-Jie Huang, and Yaqian Li.
\newblock u-llava: Unifying multi-modal tasks via large language model.
\newblock \emph{arXiv preprint arXiv:2311.05348}, 2023.

\bibitem[Xu et~al.(2024)Xu, Zhou, Yan, Gu, Arnab, Sun, Wang, and Schmid]{pixelLLM}
Jiarui Xu, Xingyi Zhou, Shen Yan, Xiuye Gu, Anurag Arnab, Chen Sun, Xiaolong Wang, and Cordelia Schmid.
\newblock Pixel-aligned language model.
\newblock In \emph{Proceedings of the IEEE/CVF Conference on Computer Vision and Pattern Recognition}, pages 13030--13039, 2024.

\bibitem[Yan et~al.(2023)Yan, Jiang, Wu, Wang, Yuan, Luo, and Lu]{UNINEXT}
Bin Yan, Yi Jiang, Jiannan Wu, Dong Wang, Zehuan Yuan, Ping Luo, and Huchuan Lu.
\newblock Universal instance perception as object discovery and retrieval.
\newblock In \emph{CVPR}, 2023.

\bibitem[Yang et~al.(2023)Yang, Qu, Lai, Tian, Peng, Liu, and Jia]{lisa++}
Senqiao Yang, Tianyuan Qu, Xin Lai, Zhuotao Tian, Bohao Peng, Shu Liu, and Jiaya Jia.
\newblock An improved baseline for reasoning segmentation with large language model.
\newblock \emph{arXiv preprint arXiv:2312.17240}, 2023.

\bibitem[Yang et~al.(2022)Yang, Wang, Tang, Chen, Zhao, and Torr]{lavt}
Zhao Yang, Jiaqi Wang, Yansong Tang, Kai Chen, Hengshuang Zhao, and Philip~HS Torr.
\newblock Lavt: Language-aware vision transformer for referring image segmentation.
\newblock In \emph{Proceedings of the IEEE/CVF Conference on Computer Vision and Pattern Recognition}, pages 18155--18165, 2022.

\bibitem[Yao et~al.(2024)Yao, Wang, Ye, and Liu]{yao2024matte}
Jingfeng Yao, Xinggang Wang, Lang Ye, and Wenyu Liu.
\newblock Matte anything: Interactive natural image matting with segment anything model.
\newblock \emph{Image and Vision Computing}, 147:\penalty0 105067, 2024.

\bibitem[Ye et~al.(2019)Ye, Rochan, Liu, and Wang]{ye2019cross}
Linwei Ye, Mrigank Rochan, Zhi Liu, and Yang Wang.
\newblock Cross-modal self-attention network for referring image segmentation.
\newblock In \emph{Proceedings of the IEEE/CVF conference on computer vision and pattern recognition}, pages 10502--10511, 2019.

\bibitem[Yu et~al.(2016)Yu, Poirson, Yang, Berg, and Berg]{refcoco/+}
Licheng Yu, Patrick Poirson, Shan Yang, Alexander~C Berg, and Tamara~L Berg.
\newblock Modeling context in referring expressions.
\newblock In \emph{Computer Vision--ECCV 2016: 14th European Conference, Amsterdam, The Netherlands, October 11-14, 2016, Proceedings, Part II 14}, pages 69--85. Springer, 2016.

\bibitem[Yu et~al.(2023)Yu, Feng, Feng, Liu, Jin, Zeng, and Chen]{yu2023inpaint}
Tao Yu, Runseng Feng, Ruoyu Feng, Jinming Liu, Xin Jin, Wenjun Zeng, and Zhibo Chen.
\newblock Inpaint anything: Segment anything meets image inpainting.
\newblock \emph{arXiv preprint arXiv:2304.06790}, 2023.

\bibitem[Zhang et~al.(2023)Zhang, Han, Qiao, Kim, Bae, Lee, and Hong]{mobile_sam}
Chaoning Zhang, Dongshen Han, Yu Qiao, Jung~Uk Kim, Sung-Ho Bae, Seungkyu Lee, and Choong~Seon Hong.
\newblock Faster segment anything: Towards lightweight sam for mobile applications.
\newblock \emph{arXiv preprint arXiv:2306.14289}, 2023.

\bibitem[Zhang et~al.(2024)Zhang, Ma, Zhang, and Bai]{psalm}
Zheng Zhang, Yeyao Ma, Enming Zhang, and Xiang Bai.
\newblock Psalm: Pixelwise segmentation with large multi-modal model.
\newblock \emph{arXiv preprint arXiv:2403.14598}, 2024.

\bibitem[Zhao et~al.(2023)Zhao, Ding, An, Du, Yu, Li, Tang, and Wang]{fast_sam}
Xu Zhao, Wenchao Ding, Yongqi An, Yinglong Du, Tao Yu, Min Li, Ming Tang, and Jinqiao Wang.
\newblock Fast segment anything, 2023.

\bibitem[Zhou et~al.(2017)Zhou, Zhao, Puig, Fidler, Barriuso, and Torralba]{ade1}
Bolei Zhou, Hang Zhao, Xavier Puig, Sanja Fidler, Adela Barriuso, and Antonio Torralba.
\newblock Scene parsing through ade20k dataset.
\newblock In \emph{Proceedings of the IEEE conference on computer vision and pattern recognition}, pages 633--641, 2017.

\bibitem[Zhou et~al.(2019)Zhou, Zhao, Puig, Xiao, Fidler, Barriuso, and Torralba]{ade2}
Bolei Zhou, Hang Zhao, Xavier Puig, Tete Xiao, Sanja Fidler, Adela Barriuso, and Antonio Torralba.
\newblock Semantic understanding of scenes through the ade20k dataset.
\newblock \emph{International Journal of Computer Vision}, 127:\penalty0 302--321, 2019.

\bibitem[Zhu et~al.(2024)Zhu, Zhu, Liu, Xu, and Peng]{miphi}
Yichen Zhu, Minjie Zhu, Ning Liu, Zhiyuan Xu, and Yaxin Peng.
\newblock Llava-phi: Efficient multi-modal assistant with small language model.
\newblock In \emph{Proceedings of the 1st International Workshop on Efficient Multimedia Computing under Limited}, pages 18--22, 2024.

\end{thebibliography}
}

\clearpage
\setcounter{page}{1}
\maketitlesupplementary


\appendix

\section{Language-guided Multi-task Segmentation.}

Due to the limitation of Referring Expression Segmentation~(RES) data in aspects of both quantity~(80k) and quality~(mask accuracy, caption diversity), employing data from other sub-tasks of segmentation~(\eg, semantic segmentation) become popular among recent works~(\eg, LISA~\citep{lisa}, u-LLaVA~\citep{u-llava}). 
If properly implemented, additional data can not only improve RES performance, but also extend the segmentation granularities of RES~(\ie, semantic-level segmentation, part-level segmentation, introduced in Sec.~\ref{sec:semanticRES} and Sec.~\ref{sec:partRES}).
However, existing works usually apply a rough manner, leading to weak or even negative influence on models' RES performance. Moreover, they are weak at other granularites of RES.

In this section, we firstly analyze the challenges of employing data from other sub-tasks of segmentation~(\eg, semantic segmentation), then we present our detailed recipe of our data strategies to building the hybrid dataset. Finally we introduce two tasks called "Semantic-level Referring Expression Segmentation" and "Part-level Referring Expression Segmentation" to evaluate the models' performance at different granularites of RES.

\begin{table*}[t]
    \centering
    \small
    \caption{\textbf{Explanation to semantic conflict.}
        The top row contains the source image and annotated mask in the ADE20K~\citep{ade1,ade2} dataset. 
        The middle row presents the comparison of templates used by different works to apply semantic data into Referring Expression Segmentation, where \textcolor{red}{red} fonts represent the wrong description, while the \textcolor{green}{green} fonts represent the right description. 
        The bottom row contains ideal caption. 
        Previous works either prompt with the raw category name without modifications or apply a fixed template, leading to unaligned descriptions in some cases.
        We use special token `[semantic]' as the task identifier, overcoming the misalignment problem.}
    \vspace{-5pt}
    \scalebox{1}{
    \begin{tabular}{lccc}\toprule
        Image \& mask & \includegraphics[height=4cm]{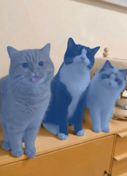} & \includegraphics[height=4cm]{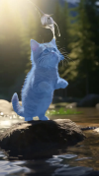} & \includegraphics[height=4cm]{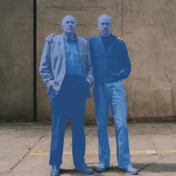} \\\midrule
        Raw~\citep{lisa} & \textcolor{red}{cat} & \textcolor{green}{cat} & \textcolor{red}{man}\\
        Fixed template~\citep{uniseg} & \textcolor{green}{all the cats} & \textcolor{red}{all the cats} & \textcolor{red}{all the mans}\\
        Special token~(ours) & \textcolor{green}{[semantic] cat} & \textcolor{green}{[semantic] cat}& \textcolor{green}{[semantic] man}\\\midrule
        Ideal caption & all the cats/cats/... & the cat/cat/... & all the men/... \\\bottomrule
    \end{tabular}
    }
    \label{tab:semantic_conflict}
    \vspace{-10pt}
\end{table*}

\subsection{Challenges.}
\label{seg:data_challenge}
Existing segmentation data can be generally separated to the semantic-level data and the instance-level data. When applying these data for Referring Expression Segmentation~(RES), Semantic conflict will arise in the semantic-data and ambiguity problem will arise in the instance-level data.

\mypara{Semantic conflict:}
The semantic conflict, as discussed in UniLSeg~\citep{uniseg}, mainly exhibits between semantic segmentation datasets and RES datasets. 
Tab.~\ref{tab:semantic_conflict} presents several representative cases. 
The semantic segmentation datasets only provide the category name as annotation, letting go of the information regarding to instance number. 
However, in RES tasks, the description should explicitly align with the related region. 
As can be seen in the first case in Tab.~\ref{tab:semantic_conflict}, `cat' does not align with all three cats in the image. 
Some works~(\ie, \citep{uniseg}) propose using fixed template like `all the {category}' to avoid the semantic conflict, while introducing additional misalignment. The over-description problem can be seen in the second case in Tab.~\ref{tab:semantic_conflict}. `All the cats' is used to describe the single cat in the image. 
Furthermore, difficulty of handling plural forms can be seen in the third case in Tab.~\ref{tab:semantic_conflict}. The fixed template would fail at categories like `bread', `man'. 
The experiment in Sec.~\ref{extra semantic data} shows how semantic conflict affects performance. 

\mypara{Ambiguity problem:}
The ambiguity problem mainly exhibits between instance-level dataset and RES datasets. 
In instance-level datasets, multiple objects may align to the same category.
However, in RES datasets, the correspondence is one-to-one. 
Directly using instance segmentation data to train an RES model will result in inconsistent target assignments, leading to unstable training.



\subsection{Unified training with multi-task datasets.}
\label{seg:data_strategy}
To solve the problems mentioned above, we propose data strategies for various datasets. Specifically, we propose filter/merging principles for instance-level data and introduce special token `[semantic]' to semantic-level data. We will open-source related codes on our project page.

\mypara{Instance-level data:}
We include Objects365~\citep{shao2019objects365} to extend RES data. Specifically, (a) for each image, we exclude categories with more than one instance to avoid ambiguity problems. (b) we employ SAM-2~\citep{ravi2024sam2} to automatically annotate masks according to the selected ground-truth bounding boxes.
Despite filtering, a substantial number of annotations remain, owing to the high density of annotations in Objects365~\citep{shao2019objects365}. 
From the original 600K images and 10M annotations, we retain 524K images and 1.8M annotations. 
The mask quality from automatic annotation is high due to the accurate ground-truth from Objects365~\citep{shao2019objects365} and the powerful segmentation capability of SAM-2~\citep{ravi2024sam2}. 
Notably, the remaining annotations are valuable for addressing long-tail problems because those excluded annotations predominantly represent head categories.

\mypara{Semantic-level data:}
We introduce ADE20K~\citep{ade1,ade2} to broaden RES models' capability of various granularities. 
We introduce a special token `[semantic]' and input `\textit{[semantic] \{category\}}' to RES models. 
The special token would not be limited to common grammar so it is helpful to avoid semantic conflict. 
As shown in Tab.~\ref{tab:semantic_conflict}, the reconstructed descriptions with our special token have no conflict with the mask annotation.

\mypara{Part-level data:}
To enable the model to segment parts of objects, we introduce PartImageNet~\citep{he2022partimagenet}, HumanParsing~\citep{liang2015deep, liang2015human} and PASCAL-Part~\citep{pascal_part} to train our model. 
For datasets annotated in semantic-level, \ie,  HumanParsing, we implement the same strategy as ADE20K. Exceptionally, we align the definition of `left' and `right' with RES datasets~(\eg, RefCOCO).
For datasets annotated in instance-level, \ie, PartImageNet and PASCAL-Part, we merge instance masks of the same category to convert the dataset to semantic-level. Then, the same strategy as ADE20K is implemented. 

By combining those datasets, we observe a significant performance gain of 1.0 cIoU in average on RefCOCO/+/g~\citep{referit,refcoco/+,refcocog,refcocog2}, as shown in Tab.~\ref{tab:benchmark}. 
Moreover, our model is empowered the capability of RES of various granularities~(\ie, semantic-level segmentation, part-level segmentation, introduced in Sec.~\ref{sec:semanticRES} and Sec.~\ref{sec:partRES}).

\subsection{Semantic-level RES}
\label{sec:semanticRES}
In this section, we will make a detailed explanation of our proposed new task `Semantic-level Referring Expression Segmentation', present the metric comparison, and provide some visualizations.

\mypara{Definition.}
Basic RES tasks only focus on the instance-level segmentation. A special prompt only refers to a single object in the image. Trained only on RefCOCO/+/g~\citep{referit,refcoco/+,refcocog,refcocog2} datasets, the model cannot correctly respond to more complex prompts~(\eg, multiple objects, stuff categories).

To this end, we propose `Semantic-level RES', aiming at improving RES models' capability of segmenting semantic-level objects.
We make perfect use of existing semantic segmentation datasets. The annotated category is managed as text prompt of the RES model, while the semantic mask corresponding to the category is managed as prediction target.
Notably, we include stuff categories during training and evaluation. 

`Semantic-level RES' differs from semantic segmentation in that we manage object category as model input, not the output.
`Semantic-level RES' differs from open vocabulary segmentation in that this is an in-domain task.

\mypara{Metrics on Semantic-level RES. }
Being a sub-task of Referring Segmentation, we take cIoU and gIoU as metrics following the RES setting. This metric examines the competence of a model to segment multi-objects and stuff categories.

We compare our model with LISA~\citep{lisa} and u-LLaVA~\citep{u-llava}, both of which also include ADE20K~\citep{ade1, ade2} data during training. 
As shown in Tab.~\ref{tab:semantic-level}, our model demonstrates a significant advantage over previous methods in semantic-level RES. 
This superior performance is attributed to our novel approach to resolving the semantic conflict (\ie, using the special token `[semantic]'). 
The solution of semantic conflict greatly improves model's capability of segmenting background classes~(\eg, `sky', `sea', or simply `background') and segmenting multi-objects~(\eg, `people', `flowers').

\begin{figure*}[t]
    \centering
    \includegraphics[width=0.7\linewidth]{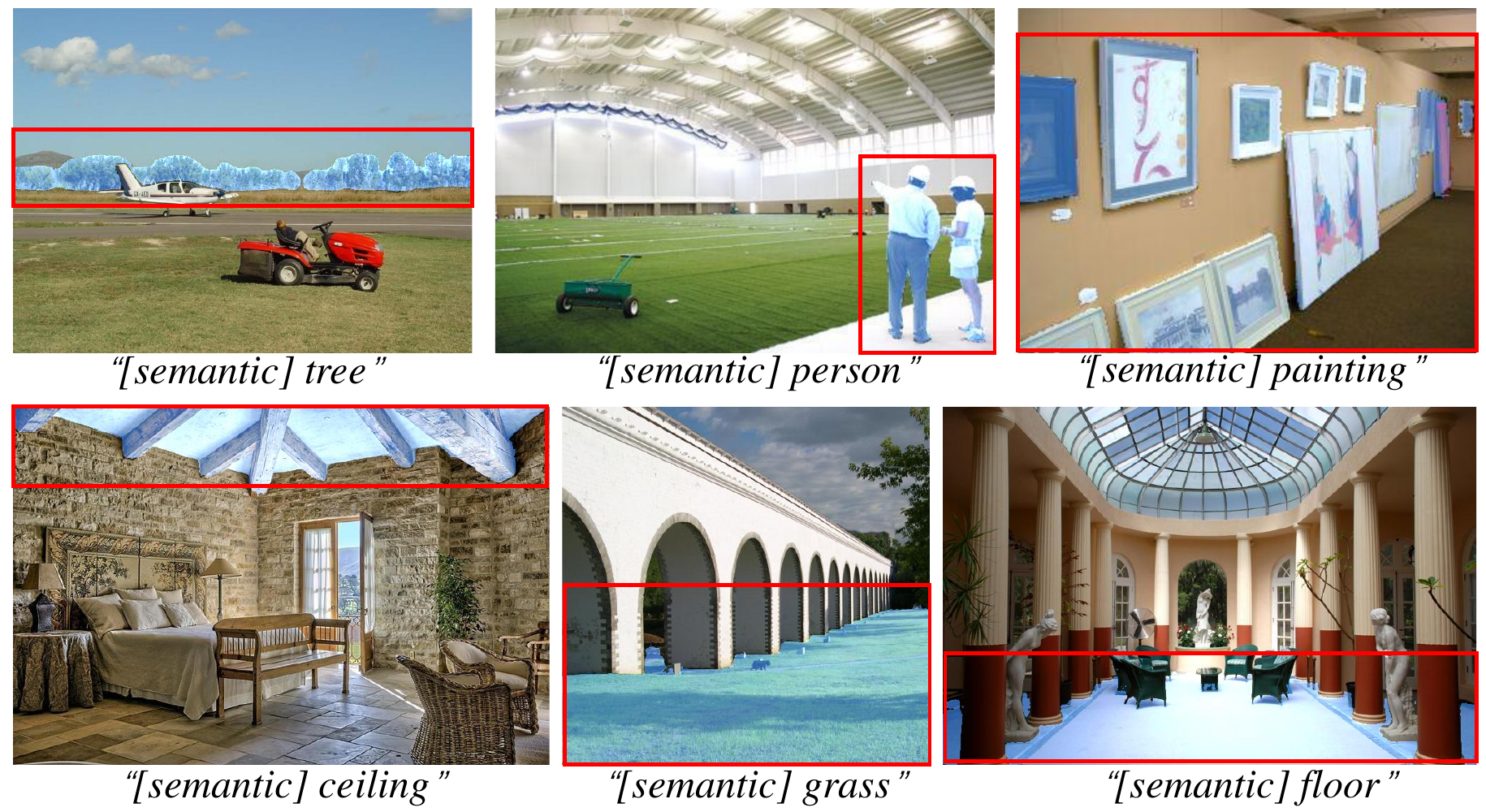}\vspace{-7pt}
    \caption{\textbf{Visualization Results on ADE20K Val.} We add a special token `[semantic]' in front of category name to construct the referring prompt, as written below each image. We use red boxes to highlight the targeted object and use blue masks to visualize the segmentation result. Upper three images showcase \name{}'s capability at multi-object RES. Below three images showcase \name{}'s capability at stuff-object RES.}
    \vspace{-10pt}
    \label{fig:vis_ade20k}
\end{figure*}

\mypara{Visualizations on ADE20K.}
Fig.~\ref{fig:vis_ade20k} shows the visualization of `Semantic-level Referring Segmentation' task on ADE20K val. Given an semantic annotated category as text prompt for RES models, the models are to predict semantic masks containing all pixels corresponding to the category. Visualization results are provided by our \name{}. Trained on Semantic-level Referring Segmentation tasks, \name{} is able to segment multiple objects~(\eg, drawings, chairs) or background categories~(\eg, floor, ground).

\begin{table}[t]
    \centering
    \small
    \caption{\textbf{Semantic-level Referring Segmentation results.} * denotes zero-shot results. \name{} outperforms previous methods that include ADE20K~\citep{ade1,ade2} during training, indicating the effectiveness of our proposed special token `[semantic]'.}
    \vspace{-5pt}
    \setlength{\tabcolsep}{7pt}
    \scalebox{0.8}{
    \begin{tabular}{lcccc}\toprule
        \multirow{2}{*}{Method} & \multicolumn{2}{c}{ADE20K~\citep{ade1,ade2}} & \multicolumn{2}{c}{Pascal VOC~\citep{everingham2010pascal}}\\\cmidrule{2-5}
        &gIou & cIoU & gIou & cIoU\\\midrule
        LISA-7B & 49.0 & 67.4 & 53.6* & 46.7*\\
        LISA-7B-explanatory & 55.4 & 70.8 & 56.2* & 47.0*\\
        u-LLaVA & 42.1 & 75.3 & 57.5 & 44.8 \\
        EVF-SAM & \textbf{64.5} & \textbf{76.8} & \textbf{78.2*} & \textbf{81.8*}\\\bottomrule
    \end{tabular}
    }
    \vspace{-20pt}
    \label{tab:semantic-level}
\end{table}


\subsection{Part-level RES}
\label{sec:partRES}

In this section, we will make a detailed explanation of our proposed new task `Part-level RES', present the metric comparison, and provide some visualizations.

\mypara{Definition}
Similar to Semantic-level RES, we further propose `Part-level RES', aiming at improving RES models' capability of segmenting part of objects.
We make perfect use of existing part segmentation datasets. The annotated category is managed as text prompt of the RES model, while the semantic mask corresponding to the category is managed as prediction target.

\mypara{Metrics on Part-level RES}
We evaluate part-level RES on Pascal-Part dataset.
It is worth noting that we merge the instance-level annotations to semantic-level annotations to avoid ambiguity problems, as illustrated in Sec.~\ref{data strategy}.
We still following the RES settings to evaluate cIoU and gIoU as validation metrics.
The metrics are shown in Tab.~\ref{tab:part-level}. We compare our \name{} with LISA~\citep{lisa}, which also includes Pascal-Part~\citep{pascal_part} dataset during training.
The experiment results show that our \name{} outperforms LISA at part-level RES task by a huge margin.

\begin{figure*}[t]
    \centering
    \includegraphics[width=0.7\linewidth]{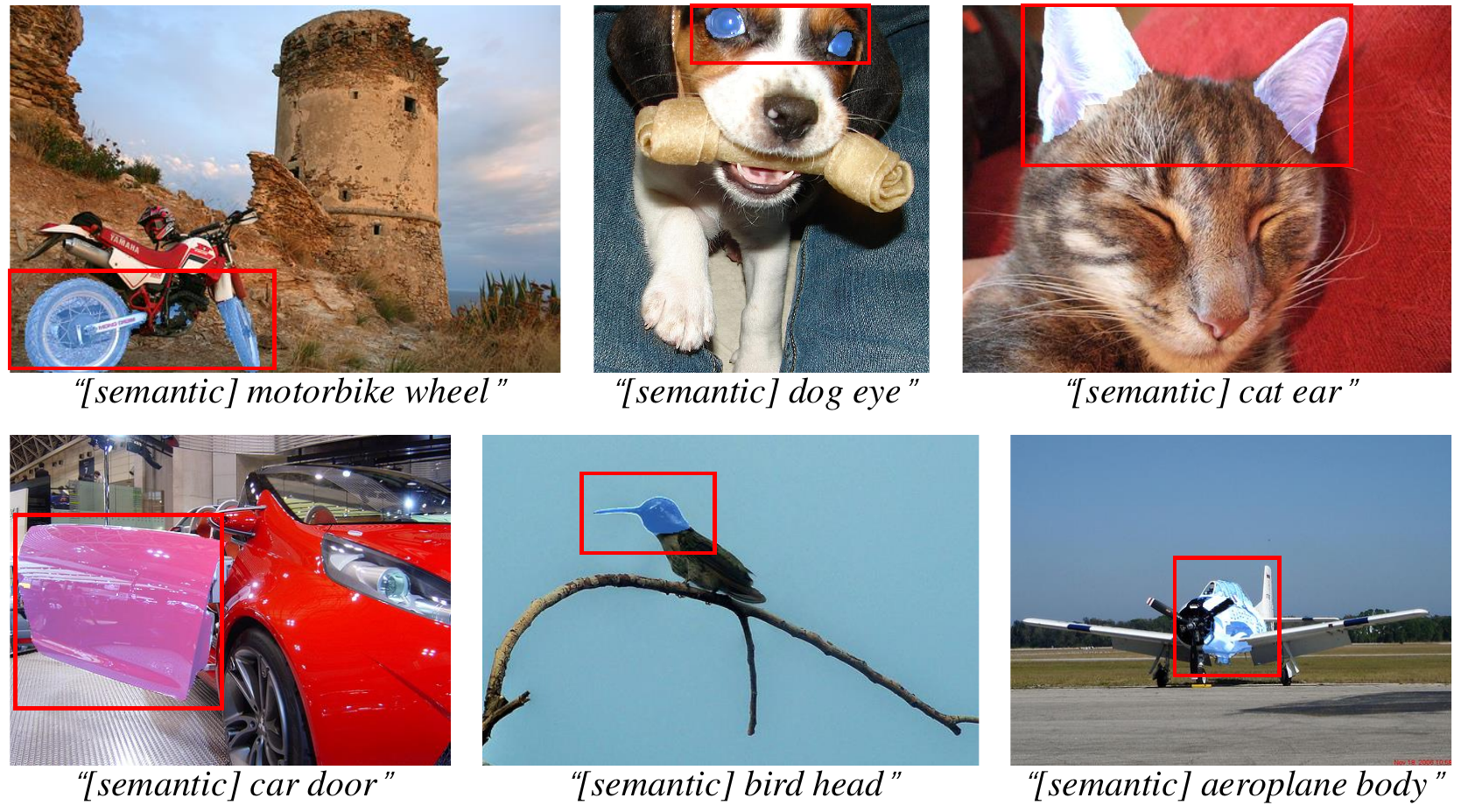}
    \vspace{-5pt}
    \caption{\textbf{Visualization Results on Pascal-Part Val.} We add a special token `[semantic]' in front of category name to construct the referring prompt, as written below each image. We use red boxes to highlight the targeted object and use blue masks to visualize the segmentation result. Upper three images showcase \name{}'s capability at multi-part RES. Below three images showcase \name{}'s capability at broad-part RES.}
    \vspace{-10pt}
    \label{fig:vis_pascal_part}
\end{figure*}

\mypara{Visualizations on Pascal-Part.}
Fig.~\ref{fig:vis_pascal_part} shows the visualization of `Part-level Referring Expression Segmentation' on Pascal-Part val. 

\begin{table}[t]
    \centering
    \small
    \caption{\textbf{Part-level Referring Expression Segmentation results.} 
    We present evaluation metrics on Pascal-Part and compare the results with LISA, which also includes Pascal-Part during training.
    Our \name{} outperforms LISA by a large margin.}
    \vspace{-5pt}
    \begin{tabular}{ccc}\toprule
        Method & gIoU & cIoU\\\midrule
        LISA & 21.2 & 27.0\\
        \name{} & \textbf{53.7} & \textbf{71.4}\\\bottomrule
    \end{tabular}
    \vspace{-15pt}
    \label{tab:part-level}
\end{table}

\section{Additional Experiments}

\subsection{Reasoning segmentation.}
\label{sec:reasonSeg}
ReasonSeg~\citep{lisa} is a Referring Expression Segmentation dataset targeting long and indirect descriptions. 
It comprises 239 training images, 200 validation images and 779 test images. 
Text annotations of ReasonSeg are posed in an instructional format, such as `What is the food with the most Vitamin C in this image?'. 
Considering the limited amount of data and the risk of over-fitting, we process evaluation using both the zero-shot setting and the finetuning setting. 
The zero-shot performance on ReasonSeg of our \name{} surpasses LLaVA-free methods~(\eg, OVSeg, SEEM), but falls behind methods with LLaVA1.1~\citep{llava}.
When we finetune our \name{} with 239 ReasonSeg training samples, we observe significant improvement in reasoning performance, outperforming LISA-7B-LLaVA1.1~\citep{lisa}
However, finetuned \name{} still falls behind LISA with LLaVA1.5~\citep{llava1.5}. 

We analyze LISA's superior performance on ReasonSeg compared to \name{}, attributing it to several key factors:
\begin{itemize}
    \item LISA's reasoning abilities significantly improved when transitioning from LLaVA 1.1 to LLaVA 1.5. This enhancement stems from LLaVA 1.5's incorporation of reasoning-focused Visual Question Answering (VQA) datasets (e.g., CLEVR \citep{johnson2017clevr}, GQA \citep{hudson2019gqa}), which effectively transfer reasoning knowledge to ReasonSeg. 
    However, \name{}'s foundation model~(\ie, BEIT-3~\citep{beit3}) is only pre-trained using Masked Language Modeling~(MLM) recipe.
    \item LLM-free methods~(\eg, \name{}) are absent of instruction tuning, hindering their ability to accurately interpret the QA formatted data.

\end{itemize}

\begin{table}[t]
\centering
\small
\setlength{\tabcolsep}{8pt}
\caption{\textbf{Comparision of segmentation capability on ReasonSeg~\citep{lisa} dataset.} Our \name{} outperforms previous methods on both cIoU and gIoU metrics.}
\vspace{-5pt}
\scalebox{0.85}{
    \begin{tabular}{lccccc}\toprule
        \multirow{2}{*}{Method} & \multirow{2}{*}{Finetune?} & \multicolumn{2}{c}{val} & \multicolumn{2}{c}{test}\\
        &&gIoU & cIoU&gIoU & cIoU\\\midrule
        OVSeg & \xmark & 28.5 & 18.6 & 26.1 & 20.8 \\
        X-Decoder & \xmark  & 22.6 & 17.9& 21.7 & 16.3\\
        SEEM & \xmark & 25.5 & 21.2& 24.3 & 18.7 \\
        LISA-7B-LLaVA1.1 & \xmark & 44.4 & 46.0 & 36.8 & 34.1    \\
        LISA-7B-LLaVA1.1 & \checkmark  & 52.9 &  54.0& 36.8 & 34.1\\
        LISA-7B-LLaVA1.5 & \xmark & 53.6 & 52.3 & 48.7& 48.8\\
        LISA-7B-LLaVA1.5 & \checkmark & 61.3& 62.9 & 55.6 & 56.9\\
        LISA-13B-LLaVA1.5 & \checkmark & 65.0 &72.9 &61.3& 62.2\\
        EVF-SAM & \xmark & 38.3 & 30.0 & 38.9 & 32.6\\
        EVF-SAM & \checkmark & 58.7 &55.7&55.1 & 50.5\\\bottomrule
    \end{tabular}
    }
\vspace{-15pt}
\end{table}

\subsection{Experiments of simply adding extra data.}
\label{extra semantic data}
We tried to introduce some extra semantic segmentation datasets~(ADE20K~\citep{ade1,ade2}, Mapillary~\citep{mapillary}) to proceed with joint training. We do not include COCO-Stuff~\citep{cocostuff} to avoid data leakage with RefCOCO/+/g. We report comparison results in Tab.~\ref{tab:dataset}. 
From the experimental results: 
(a) We find that several metrics gain when introducing extra semantic segmentation data~(\eg, RefCOCO+/testB). This may results from the analogous data composition between RefCOCO+ and ADE20K, indicating the feasibility of introducing data of other segmentation sub-task to boost RES model. 
(b) We find that the Semantic-level Referring Segmentation capability improves a lot when including extra semantic segmentation data, indicating the necessity of building multi-task training.
(c) However, We observe performance degradation on various metrics~(\eg, RefCOCO/val, RefCOCO/testA). This is due to the difference distribution between datasets from RES and other segmentation sub-tasks.

\begin{table*}[t]
    \centering
    \small
    \setlength{\abovecaptionskip}{0.2cm}
    \caption{\textbf{Results of adding extra semantic data.} \(^*\) means zero-shot results. The reported ADE20K results are evaluated on Semantic-level Referring Segmentation tasks proposed in main text.}
    \vspace{2pt}
    \scalebox{0.9}{
    \begin{tabular}{ccccccccccc}\toprule
        \multirow{2}{*}{ADE20K} & \multirow{2}{*}{Mapillary} & \multicolumn{3}{c}{RefCOCO} & \multicolumn{3}{c}{RefCOCO+} & \multicolumn{2}{c}{RefCOCOg} & ADE20K \\\cmidrule{3-11}
        & & val & testA & testB & val & testA & testB & val & test & val\\
        \midrule
        & & \textbf{82.1} & \textbf{83.7} & 80.0 & 75.2 & 78.3 & 70.1 & \textbf{76.8} & 77.4 & ~~54.2\(^*\) \\
        \checkmark &  & 81.7 & 83.6 & \textbf{80.3} & 75.4& \textbf{78.4}& \textbf{71.3} & 75.5 &\textbf{77.6}& 75.9\\
         & \checkmark & 81.9 & 83.5 & \textbf{80.3} & 75.1 & 78.0 & 70.8 & 75.3 & 77.4 & ~~59.6\(^*\) \\
         \checkmark & \checkmark & 81.8 & 83.4 & 79.7 & \textbf{75.6} & 78.0 & 70.7 & 75.8 & 76.9 & \textbf{76.1}\\
         \bottomrule
    \end{tabular}
    }
    \vspace{-5pt}
    \label{tab:dataset}
\end{table*}

\subsection{Multimodal feature representation.}
\begin{table}[h]
\centering
\small
\caption{\textbf{Ablations on multimodal feature representation.} BEIT-3 contains two \texttt{[CLS]} tokens for visual and textual modalities. We also explore the effects of \texttt{AvgPool} and late fusion between two modalities.}
\vspace{-10pt}
\setlength{\tabcolsep}{3.5pt}
\scalebox{1}{
\begin{tabular}{ccccc}\toprule
    \makecell{\texttt{[CLS]}$_{\text{Text}}$} & \makecell{\texttt{[CLS]}$_{\text{Image}}$} &
    \makecell{\texttt{AvgPool}$_{\text{Image}}$}
    &\makecell{Fusion} & \makecell{cIoU}\\\midrule
    \checkmark & & & - & 83.5   \\
    & \checkmark & & - & \textbf{83.7}\\
     & & \checkmark & - & 83.5\\
     \checkmark & \checkmark & & Concat & 83.2 \\
     \bottomrule
\end{tabular}}
\vspace{-15pt}
\label{tab:representation}
\end{table}

In Tab.~\ref{tab:representation}, we explore the effects of using different multimodal features representations as prompts for SAM.
Specifically, we adopt different outputs of the Multimodal Encoder: (a) the image \texttt{[CLS]} token, (b) the \texttt{AvgPool} over image tokens, and (c) the text \texttt{[CLS]} token.
Tab.~\ref{tab:representation} shows that using image \texttt{[CLS]} token is more effective while combining image and text tokens through concatenation leads to a performance drop.

\begin{figure*}[t]
    \centering
    \includegraphics[width=0.9\linewidth]{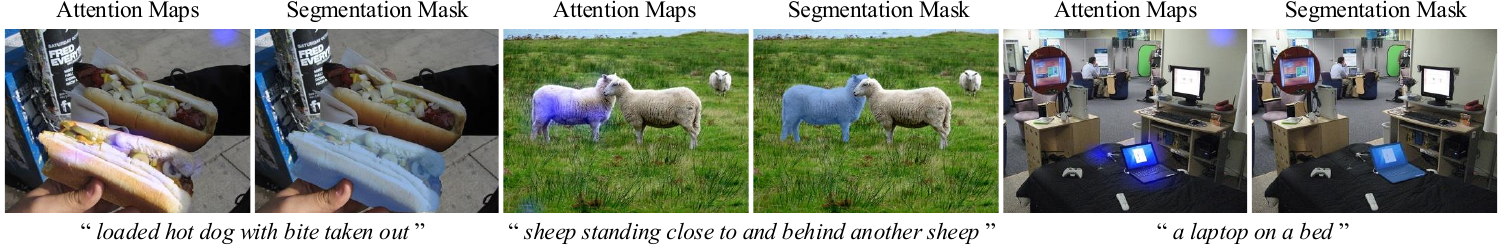}
    \vspace{-7pt}
    \caption{\textbf{Visualizations of Attention Maps in Multimodal Encoder.} To unveil the effects of the Multimodal Encoder, we visualize the attention maps between the \texttt{[CLS]} token and image tokens in the last layer of BEIT-3-Large. Specifically, we sum up the attention maps from all heads.}
    \vspace{-10pt}
    \label{fig:encoder_attn_map}
\end{figure*}

To unveil how the multimodal encoder contributes to prompting SAM with texts, we visualize the attention maps between the \texttt{[CLS]} token (prompt embeddings) and the image tokens from the last layer of BEIT-3. 
As shown in Fig.~\ref{fig:encoder_attn_map}, the attention maps focus on the target objects and are consistent with the input text prompts.
The deep fusion of text and image embeddings leads to accurate region-text alignment. 
Consequently, the prompt embeddings contain abundant object-related information, including semantics and spatial localization, which is conducive to SAM achieving precise object segmentation.

\section{Qualitative Results}
\begin{figure*}[t]
    \centering
    \includegraphics[width=0.8\linewidth]{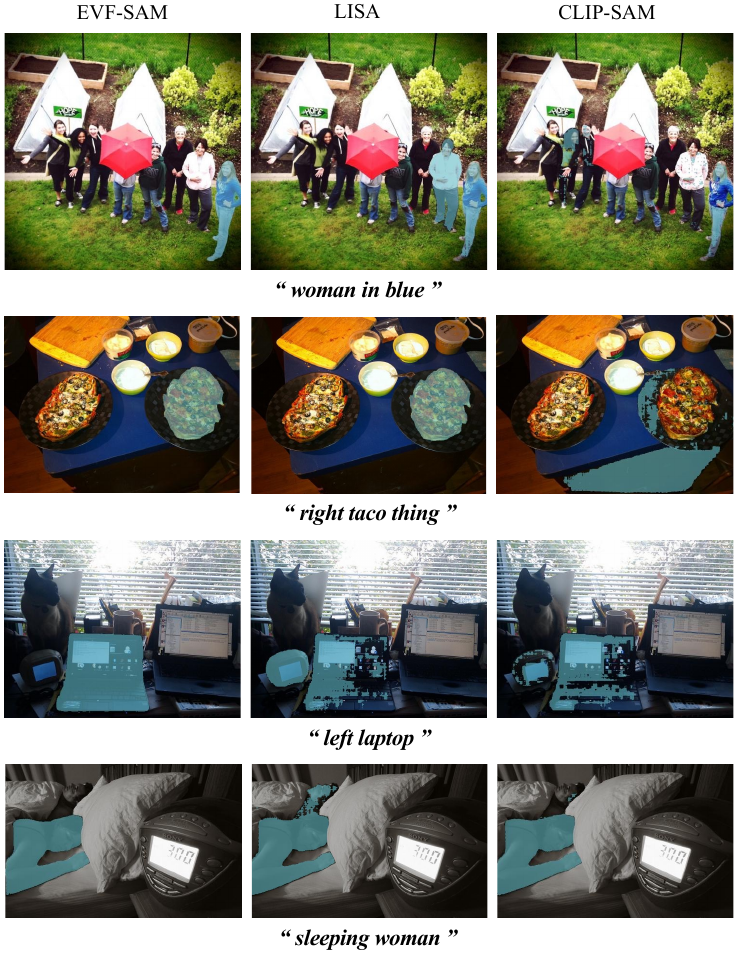}
    \caption{\textbf{Visualization Results on RefCOCO val.} We compared the qualitative results on RefCOCO which contains simple descriptive expressions.}
    \label{fig:vis_refcoco}
\end{figure*}

\begin{figure*}[t]
    \centering
    \includegraphics[width=0.8\linewidth]{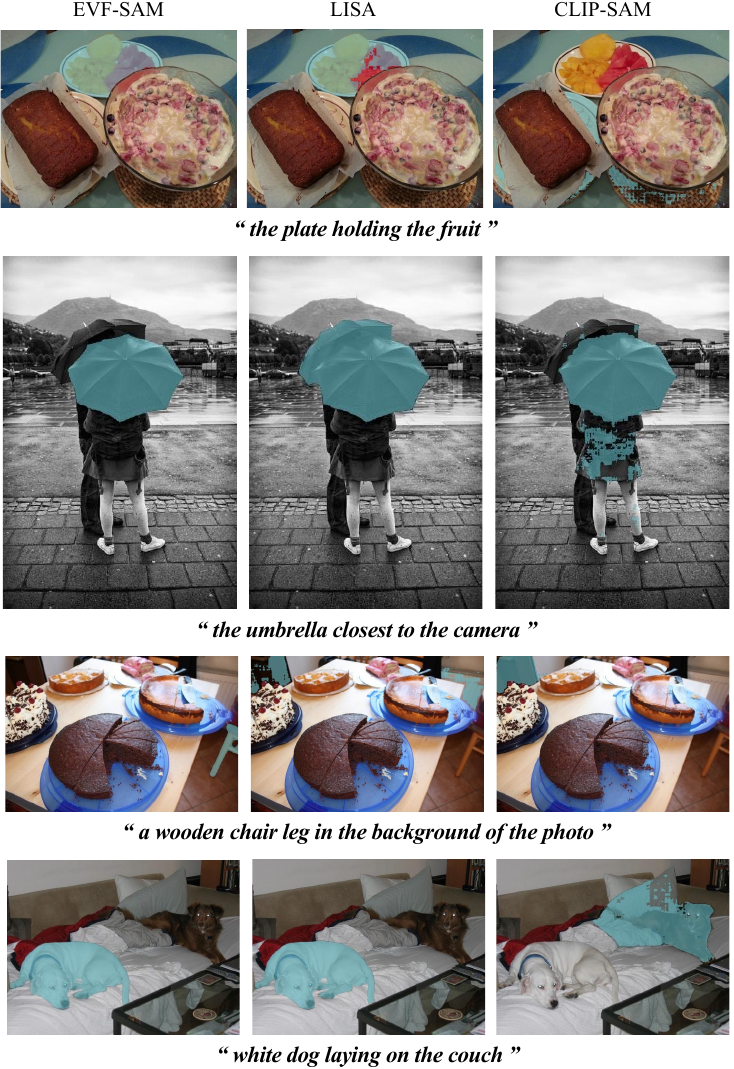}
    \caption{\textbf{Visualization Results on RefCOCOg val.} Considering that RefCOCOg contains longer expressions and we provide qualitative results to show the capability of our \name{} for understanding long expressions.}
    \label{fig:vis_refcocog}
\end{figure*}

In this section, we mainly visualize the qualitative results on  RefCOCO \textit{val} and RefCOCOg \textit{val} datasets, as shown in Fig.~\ref{fig:vis_refcoco} and Fig.~\ref{fig:vis_refcocog}, respectively.
Moreover, we compare the qualitative results of different ways to prompt SAM with texts: (1) our proposed \name{}, (2) SAM with LLM (LISA~\citep{lisa}), and (3) SAM with a CLIP text encoder implemented in this paper (suggested by \citep{sam}, which are based on the same SAM-Huge model.
The qualitative results can demonstrate the superiority of the proposed \name{}.

\mypara{Visualizations on RefCOCO.}
Fig.~\ref{fig:vis_refcoco} shows the qualitative comparisons on the RefCOCO \textit{val}, which contains simple \textit{descriptive} expression texts.
The proposed \name{} can follow the expressions and segment more accurately with clear boundaries.

\mypara{Visualizations on RefCOCOg.}
Fig.~\ref{fig:vis_refcocog} illustrates the qualitative comparisons on the RefCOCOg \textit{val}, which aims to segment objects with \textit{long} expression texts.
The SAM with a vanilla CLIP text encoder produces inferior segmentation results given the long-expression texts. However, the proposed \name{} outperforms LISA when using long expressions, even though LISA 
adopts LLaMA-7B~\citep{llama} to understand the instructions and generate prompt embeddings, showcasing that the lightweight vision-language models can understand complex expressions.
In addition, the proposed \name{} can also understand the texts or expressions towards spatial locations, such as \textit{`the umbrella closest to the camera'}.

\end{document}